\documentclass[letterpaper, 10 pt, conference]{ieeeconf}

\IEEEoverridecommandlockouts                              
                                                          
\usepackage{times}
\usepackage{amsmath}
\usepackage{amssymb}
\usepackage{graphics}
\usepackage{epsfig}
\usepackage[tight]{subfigure}

\usepackage{tabularx}
\usepackage{booktabs}
\usepackage{multirow}

\usepackage{xspace} 

\newcommand{\ie}{{i.e.,}\xspace}
\newcommand{\eg}{{e.g.,}\xspace}

\usepackage{flushend}

\usepackage{color}
\usepackage[textsize=scriptsize]{todonotes} 
\makeatletter
\let\NAT@parse\undefined
\makeatother

\usepackage[numbers,sort&compress]{natbib}
\bibpunct{[}{]}{,}{n}{}{;}

\title{\LARGE \bf
  Jointly Optimizing Placement and Inference\\ for Beacon-based Localization
}

\author{Charles Schaff \qquad David Yunis \qquad Ayan Chakrabarti \qquad Matthew R.\ Walter%
\thanks{C.\ Schaff, A.\ Chakrabarti, and M.R.\ Walter are with the
    Toyota Technological Institute at Chicago, Chicago, IL USA
    {\tt\footnotesize \{cbschaff, ayanc, mwalter\}@ttic.edu}.}%
  \thanks{D.\ Yunis is with the University of Chicago, Chicago, IL USA
    {\tt\footnotesize dyunis@uchicago.edu}.}%
}

\usepackage[hidelinks]{hyperref}

\makeatletter
\hypersetup{pdftitle={\@title},pdfauthor={Charles Schaff and David Yunis and Ayan Chahrabarti and Matthew R. Walter}}
\makeatother

\begin{document}

\maketitle

\begin{abstract}
  The ability of robots to estimate their location is crucial for a
  wide variety of autonomous operations. In settings where GPS is
  unavailable, measurements of transmissions from fixed beacons
  provide an effective means of estimating a robot's location as it
  navigates. The accuracy of such a beacon-based localization system
  depends both on how beacons are distributed in the environment, and
  how the robot's location is inferred based on noisy and potentially
  ambiguous measurements. We propose an approach for making these
  design decisions automatically and without expert supervision, by
  explicitly searching for the placement and inference strategies that,
  together, are optimal for a given environment.  Since this search is
  computationally expensive, our approach encodes beacon placement as
  a differential neural layer that interfaces with a neural network
  for inference. This formulation allows us to employ standard
  techniques for training neural networks to carry out the joint
  optimization.  We evaluate this approach on a variety of
  environments and settings, and find that it is able to discover
  designs that enable high localization accuracy.
\end{abstract}

\section{Introduction} \label{sec:introduction}

Measurements obtained through a distributed network of sensors or beacons can be an effective means of monitoring location, or the spatial distribution of other phenomena. The measurements themselves only provide indirect or noisy information towards the physical properties of interest, and so
additional computational processing is required for inference. Such inference must be designed to take into account possible degradations in the measurements, and exploit prior statistical knowledge of the environment. However, the success of inference, in the end, is limited by how the beacons and sensors were physically distributed in the first place.

Consider location-awareness, which is critical to human and robot navigation, resource discovery, asset tracking, logistical operations, and resource allocation~\cite{teller03}. In situations for which GPS is unavailable (indoors, underground, or underwater) or impractical, measurements of transmissions from fixed beacons~\cite{kurth03, newman03, olson04, djugash06, isler06, detweiler08, amundson09, huang11, kennedy12, derenick13} provide an attractive alternative. Designing a system for beacon-based location-awareness requires simultaneously deciding (a) how the beacons should be distributed (\eg spatially and across transmission channels); and (b) how location should be determined based on measurements of signals received from these beacons.

\begin{figure}[!t]
    \centering
    \includegraphics[width=\linewidth]{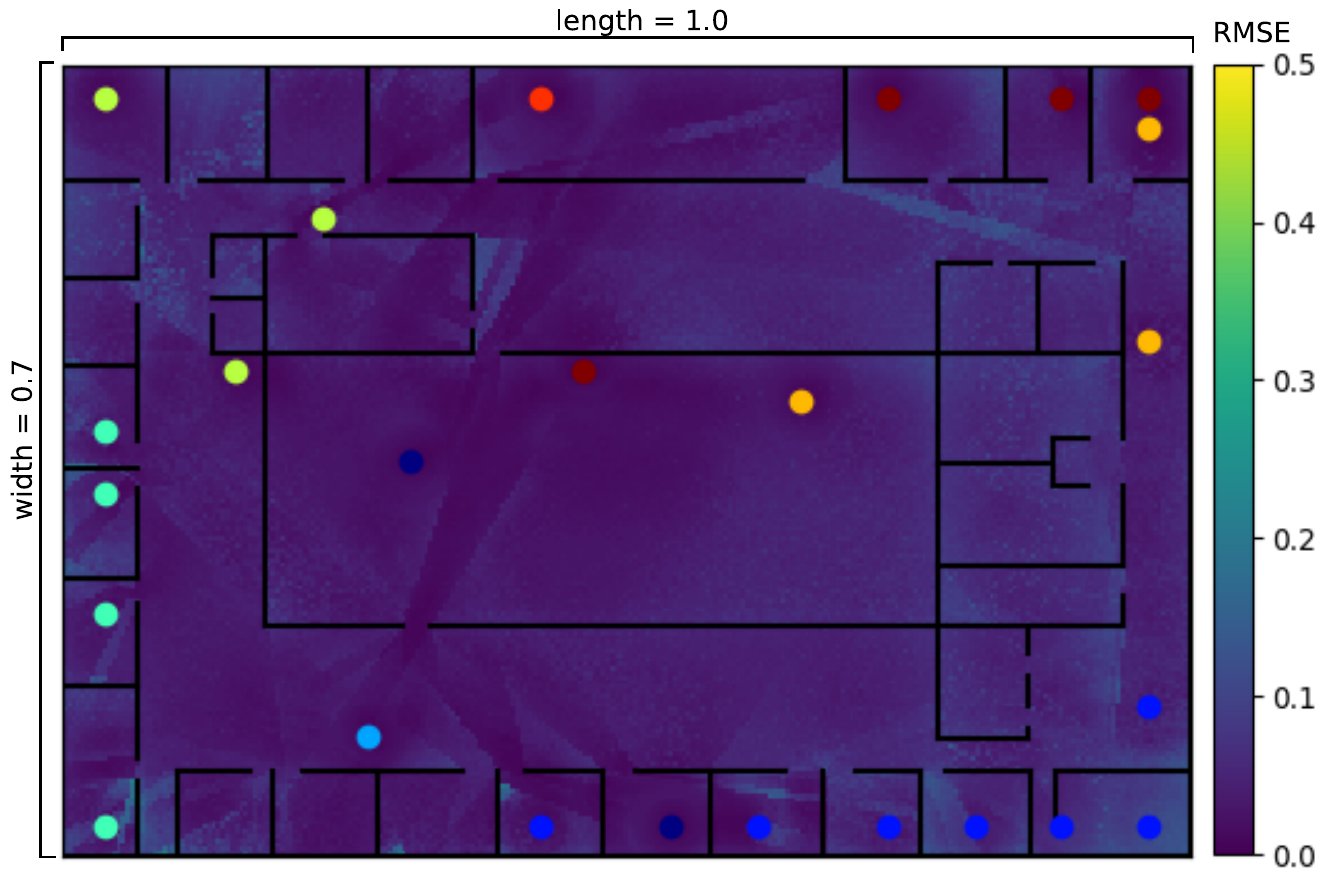}
    \caption{Our algorithm optimizes beacon placement and channel
      assignments jointly with an algorithm for location inference,
      based on measurements of transmissions from these beacons.
      receiver localization. Here, we show an example beacon
      distribution inferred by our method (colored circles denoting
      beacons and channel assignments), and the corresponding error (RMSE) in
      locations estimated by the inference method.} \label{fig:motivation}
\end{figure}

Note that these decisions are inherently coupled. The placement of beacons and their channel allocation influence the nature of the ambiguity in measurements at different locations, and therefore which inference strategy is optimal. Therefore, one should ideally search over the space of both beacon allocation---which includes the number of beacons, and their placement and channel assignment---and inference strategies to find a pair that is jointly optimal. Unfortunately, due to phenomena such as noise, interference, and attenuation due to obstructions (\eg walls), this search rarely has a closed-form solution in all but the most simplistic of settings.

Consequently, these decisions are decoupled in practice, and beacons are placed with some specific inference strategy in mind. This placement is most often performed manually by expert designers. In some cases, automatic placement methods are employed that apply heuristics (\eg coverage or field-of-view~\cite{gonzales-banos01}). However, such heuristics often rely on simplistic assumptions regarding the sensor and environment geometry, and do not adequately account for noise, interference, or other forms of degradation---\eg ignoring interference or attenuation due to walls to directly map signal strength to observations of range or bearing. These heuristics are thus unsuitable for use in many real-world settings.

Recently, Chakrabarti~\cite{chakrabarti16} introduced a method that successfully uses stochastic gradient descent (SGD) to jointly learn sensor multiplexing patterns and reconstruction methods in the context of imaging. Motivated by this, we propose a new learning-based approach to designing the beacon distribution (across space and channels) and inference algorithm \emph{jointly} for the task of localization based on raw signal transmissions. We instantiate the inference method as a neural network, and encode beacon allocation as a differentiable neural layer. We then describe an approach to jointly training the beacon layer and inference network, with respect to a given signal propagation model and environment, to enable accurate location-awareness in that environment.

We carry out evaluations under a variety of conditions---with different environment layouts, different signal propagation parameters, different numbers of transmission channels, and different desired trade-offs against the number of beacons. In all cases, we find that our approach automatically discovers designs--each different and adapted to its environment and settings---that enable high localization accuracy. Therefore, our method provides a way to consistently create optimized location-awareness systems for arbitrary environments and signal types, without expert supervision.

\section{Related Work} \label{sec:related}

Networks of sensors and beacons have been widely used for localization, tracking, and measuring other spatial phenomena. Many of the design challenges in sensor and beacon are related, since they involve problems that are duals of one another---based on whether the localization target is transmitting or receiving. In our work, we focus on localization with beacons, \ie where an agent estimates its location based on transmissions received from fixed, known landmarks.

Most of the effort in localization is typically devoted to finding accurate inference methods, assuming the distribution and location of beacons in the environment are given. One setting for these methods is where sensor measurements can be assumed to provide direct, albeit possibly noisy, measurements of relative range or bearing from beacons---an assumption that is typically based on a simple model for signal propagation. Location estimation then proceeds by using these relative range and/or bearing estimates and knowledge of beacon locations. For example, acoustic long baseline (LBL) networks are frequently used to localize underwater vehicles~\cite{newman03,olson04}, while a number of low-cost systems exist that use RF and ultrasound to measure range~\cite{priyantha05, gu09}.

\citet{moore04} propose an algorithm for estimating location based upon noise-corrupted range measurements, formulating the problem as one of realizing a two-dimensional graph whose structure is consistent with the observed ranges. \citet{detweiler08} describe a geometric technique that estimates a robot's location as it navigates a network of fixed beacons using either range or bearing observations. \citet{kennedy12} employ spectral methods to localize camera networks using relative angular measurements, and \citet{shareef08} use feed-forward and recurrent neural networks for localization based on noisy range measurements.

Another approach, common for radio frequency (RF) beacon and WiFi-based networks, is to infer location directly from received signal strength (RSS) signatures. One way to do this is by matching against a database of RSS-location pairs~\cite{prasithsangaree02}. This database is typically generated manually via a site-survey, or ``organically'' during operation~\cite{haeberlen04, park10}. \citet{sala10} and \cite{altini10} adopt a different approach, training neural networks to predict a receiver's location within an existing beacon network 
based upon received signal strength.

The above methods deal with optimal ways of inferring location given an existing network of beacons. The decision of how to distribute these beacons, however, is often made manually based on expert intuition. Automated placement methods are used rarely, and for very specific settings, such as RSS fingerprint-based localization~\cite{fang10}. The most common of these is to ensure full coverage---\ie to ensure that all locations are within an ``acceptable range'' of at least one beacon, assuming this condition is sufficient to guarantee accurate localization.

One common instance of optimizing placement for coverage is the standard art-gallery visibility problem~\cite{gonzales-banos01} that seeks placements that ensure that all locations have line-of-sight to at least one beacon. This problem assumes a polygonal environment and that the beacons have an unlimited field-of-view, subject to occlusions by walls (\eg cameras). Related, \citet{agarwal09} propose a greedy landmark-based method that solves for the placement of the minimum number of beacons (within a $\log$ factor) necessary to cover all but a fraction of a given polygonal environment. Note that these methods treat occlusions as absolute, while in practice, obstructions often only cause partial attenuation---with points that are close but obstructed observing similar signal strengths as those that are farther away. \citet{kang13} provide an interesting alternative, and like us, use backpropagation to place WiFi access points---but again, only for the objective of maximizing coverage. \citet{fang10} consider localization accuracy for placing wireless access points to maximize receiver signal-to-noise ratios.

The above methods address spatial placement but not transmission channel assignments, and associated issues with interference. Automatic channel assignment methods have been considered previously, but only for optimizing communication throughput~\cite{hills02,ling06}---\ie to minimize interference from two beacons in the same channel at any location. Note that this is a very different and simpler objective than one of enabling accurate localization, where the goal is to ensure
that there is a unique mapping from every RSS signature (with or without interference) to location.

Our approach provides a way to trade-off localization accuracy with the number of beacons, similar to the performance-cost trade-offs considered by the more general problem of sensor selection~\cite{cameron90,isler05,joshi09}. Some selection strategies are designed with specific inference strategies in mind. \citet{shewry87} and \citet{cressie93} use a greedy entropy-based approach to place sensors, tied to a Gaussian process (GP) model that is used for inference. However, this approach does not model the accuracy of the predictions at the selected locations. \citet{krause08} choose locations for a fixed sensor network that maximize mutual information using a GP model for phenomena over which inference is performed (\eg temperature). 
However, these formulations require that the phenomena be modeled as a GP, and are thus not suitable for the task of beacon-based localization.

\newcommand{\ti}{\tilde{I}}
\newcommand{\dd}{\mathcal{D}}
\newcommand{\te}{\tilde{\mathcal{E}}}
\newcommand{\ee}{\mathcal{E}}
\newcommand{\R}{\mathbb{R}}
\section{Approach} \label{sec:approach}

\begin{figure}[!t]
  \centering
  \includegraphics[width=\columnwidth]{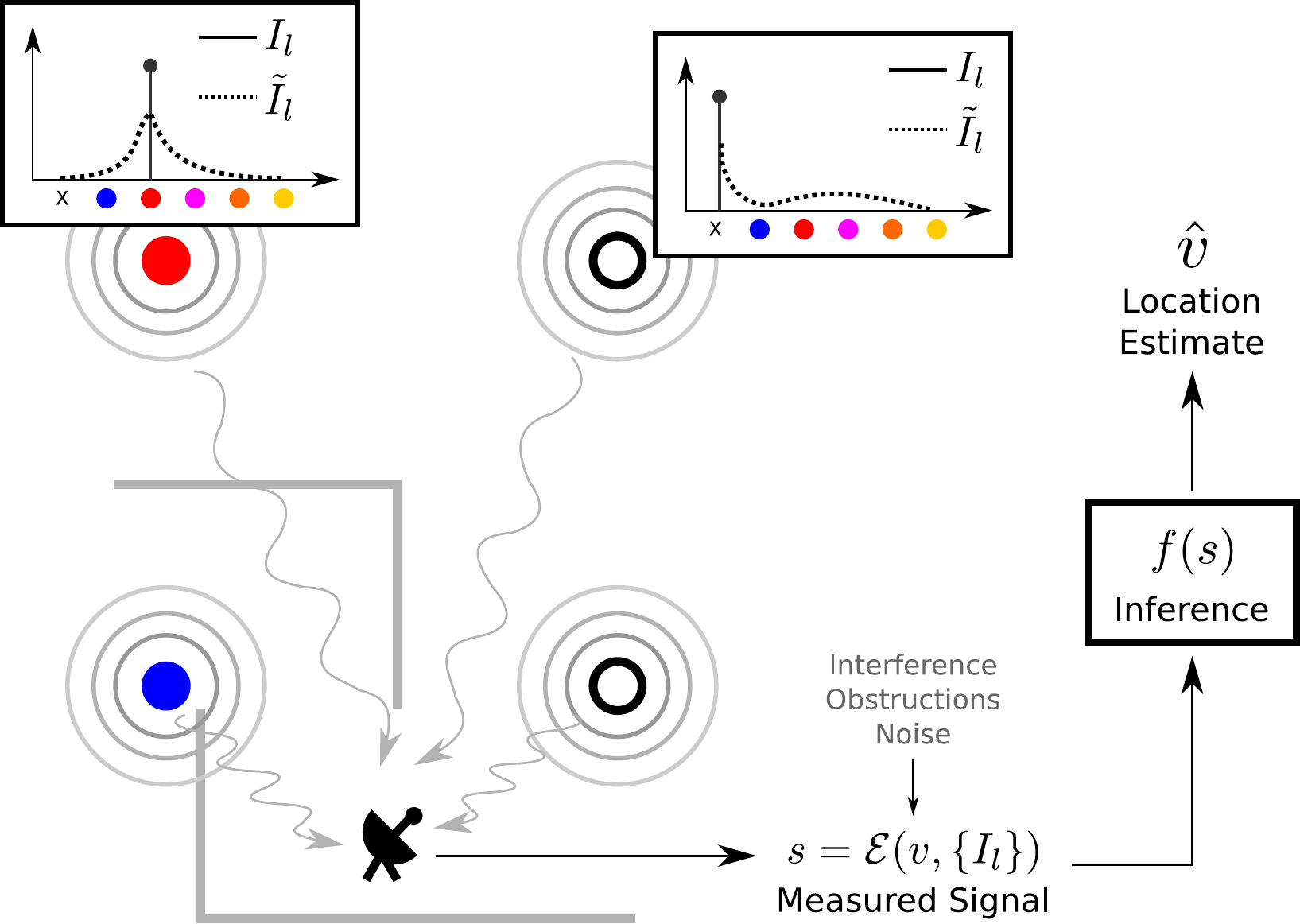}
  \caption{A schematic of our proposed approach. We seek to find the distribution of
    beacons $\{I_l\}$ and inference function $f(\cdot)$ that together
    yield reliable location estimates for a specific environment
    $\ee$. For a candidate set of locations $l$, the assignment
    vectors $I_l$ determine whether a beacon is to be placed at that
    location, and if so, in what configuration. The environment
    mapping $\ee$ then determines what the measured signal will be at
    any given location, taking into account noise, interference,
    obstructions, etc.
    We optimize these assignment vectors jointly with parameters of
    the inference function $f(\cdot)$, to minimize error in the
    estimated locations $\hat{v}$.}
  \label{fig:schema}
\end{figure}

We formalize the problem of designing a location-awareness system as
that of determining an optimal distribution of beacons $\dd$ and an
inference function $f(\cdot)$, given an environment $\ee$. For a given
set of $L$ possible locations for beacons, we parameterize the
distribution $\dd = \{I_l\}$ as an assignment $I_l$ to each location
$l \in \{1,\ldots L\}$, where $I_l\in\{0,1\}^{C+1}, |I_l|=1$ is a
$(C+1)$-dimension $0$-$1$ vector with all but one entry equal to $0$. This vector denotes whether to place a beacon at location $l$ (otherwise, $I_l^1 = 1$) and in which one of $C$ possible configurations
(i.e., $I_l^{c+1}=1$). In our experiments, a beacon's configuration corresponds to the channel on which it broadcasts.

We parameterize the environment in terms of a function $s=\ee(v,\{I_l\})$
that takes as input a location $v\in\mathbb{R}^2$ and a distribution of
beacons $\dd = \{I_l\}$, and produces a vector $s\in\R^m$ of measurements that
an agent is likely to make at that location (Fig.~\ref{fig:schema}). Note that the environment
$\ee$ need not
be a deterministic function. In the case of probabilistic phenomena such
as noise and interference, $\ee$ will produce a sample from the
distribution of possible measurements. The inference function
$f(\cdot)$ is then tasked with computing a reliable estimate of
the location given these measurements.

Our goal is to jointly optimize $\dd=\{I_l\}$ and $f(\cdot)$ such that
$f(\ee(v,\{I_l\})) \approx v$ for a distribution of possible locations
where the agent may visit. Additionally, we add a regularizer to our objective, e.g., to minimize the
total number of individual beacons.

\subsection{Optimization with Gradient Descent}

Unfortunately, the problem as stated above involves a combinatorial
search, since the space of possible beacon distributions is discrete
with $(C+1)^{L}$ elements. We make the optimization tractable by
adopting an approach similar to that of Chakrabarti~\cite{chakrabarti16}. We
relax the assignment vectors $I_l$ to be real-valued and positive as
$\ti_l \in \R_+^{C+1}, |\ti_l|=1$. The vector $\ti_l$ can be interpreted as expressing a probability distribution over the possible assignments at location
$l$.

Instead of optimizing over distribution vectors $\ti_l$ directly, we
learn a weight vector $w_l\in\R^{C+1}$ with
\begin{equation}
  \label{eq:wtoi}
  \ti_l = \mbox{SoftMax}(\alpha w_l)\qquad\ti_l^c = \frac{\exp(\alpha w_l^c)}{\sum_{c'}\exp(\alpha w_l^{c'})},
\end{equation}
where $\alpha$ is a positive scalar parameter. Since our goal is to
arrive at values of $\ti_l$ that correspond to hard assignments, we
begin with a small value of $\alpha$ and increase it during the course
of optimization according to an annealing schedule. Small values of
$\alpha$ in initial iterations allow gradients to be backpropagated
across Eqn.~\ref{eq:wtoi} to update $\{w_l\}$. As optimization
progresses, increasing $\alpha$ causes the distributions $\{I_l\}$ to
get ``peakier'', until they converge to hard assignments.

We also define a distributional version of the environment mapping
$\te(v,\{\ti_l\})$ that operates on these distributions instead of
hard assignments. This mapping can be interpreted as producing the
\emph{expectation} of the signal vector $s$ at location $v$, where the
expectation is taken over the distributions $\{\ti_l\}$. We require
that this mapping be differentiable with respect to the distribution
vectors $\{\ti_l\}$. In the next sub-section, we describe an
example of an environment mapping and its distributional version
that satisfies this requirement.

Next, we simply choose the inference function $f(\cdot)$ to be
differentiable and have some parametric form
(e.g., a neural network), and learn its parameters
jointly with the weights $\{w_l\}$ of the beacon distribution as the
minimizers of the loss:
\begin{multline}
  \label{eq:lossdef}
  L(\{w_l\},\Theta) = R(\{\ti_l\}) + \\\frac{1}{\|\mathcal{V}\|}\sum_{v\in\mathcal{V}} \mathbb{E}_s
   \left\|v-f\left(\te\left(v,\{\ti_l\}\right); \Theta \right)\right\|^2,
\end{multline}
where $\mathcal{V}$ is the set of possible agent locations, $\Theta$
are the parameters of the inference function $f$,
$\ti_l=$SoftMax$(\alpha w_l)$, as $\alpha\rightarrow\infty$, and $R$
is a regularizer. Note that the inner expectation in the second term
of Eqn.~\ref{eq:lossdef} is with respect to the distribution of possible
signal vectors for a fixed location and beacon distribution, and
captures the variance in measurements due to noise, interference, etc.

Since we require $f$ and $\te$ to be differentiable, we can optimize
both $\Theta$ and $\{w_l\}$ by minimizing Eqn.~\ref{eq:lossdef}
with stochastic gradient descent (SGD),
computing gradients over a small batch of locations
$v \in \mathcal{V}$, with a single sample of $s$ per location. We
find that the quadratic schedule for $\alpha$ used by
\citet{chakrabarti16} works well, i.e., we set
$\alpha=\alpha_0(1+\gamma t^2)$ at iteration $t$.

\subsection{Application to RF-based Localization}

To give a concrete example of an application of this framework, we
consider the following candidate setting of localization using RF
beacons. We assume that each beacon transmits a sinusoidal signal at
one of $C$ frequencies (channels). The amplitude of this signal is
assumed to be fixed for every beacon, but we allow different beacons
to have arbitrary phase variations amongst them.

We assume an agent at a location has a receiver with multiple
band-pass filters and is able to measure the power in each channel
separately (i.e., the signal vector $s$ is $C$-dimensional). We assume
that the power of each beacon's signal drops as a function of distance
and the number of obstructions (e.g., walls) in the line-of-sight
between the agent and the beacon. The measured power in each channel
at the receiver is then based on the amplitude of the super-position
of signals from all beacons transmitting on that channel. This
super-position is a source of interference, since individual beacons
have arbitrary phase. We also assume that there is some measurement
noise at the receiver.

We assume all beacons transmit at power $P_0$, and model the power of
the attenuated signal received from beacon $l$ at location $v$ as
\begin{equation}
  \label{eq:plv}
  P_l(v) = P_0~r_{l:v}^{-\zeta}~\beta^{o_{l:v}},
\end{equation}
where $\zeta$ and $\beta$ are scalar parameters, $r_{l:v}$ is the distance
between $v$ and the beacon location $l$, and $o_{l:v}$ is the number
of obstructions intersecting the line between them. The measured
power $s=\ee(v,\{I_l\})$ in each channel at the receiver is then
modeled as
\begin{multline}
  \label{eq:meas}
  s^c = \left[\epsilon_1 + \sum_l I_l^{c+1}\sqrt{P_l(v)}\cos \phi_l\right]^2 +\\
\left[\epsilon_2 + \sum_l I_l^{c+1}\sqrt{P_l(v)}\sin \phi_l\right]^2,
\end{multline}
where $\phi_l$ is the phase of beacon $l$, and
$\epsilon_1$ and $\epsilon_2$ correspond to sensor noise.  We also model
sensor saturation by clipping $s^c$ at some threshold $\tau$. At each invocation of the environment function, we randomly sample the
phases $\{\phi_l\}$ from a uniform distribution between $[0,2\pi)$,
and noise terms $\epsilon_1$ and $\epsilon_2$ from a zero-mean Gaussian distribution with
variance $\sigma_z^2$.

During training, the distributional version of the
environment function $\te$ is constructed simply by replacing $I_l$ with
$\ti_l$ in Eqn.~\ref{eq:meas}. For regularization, we use a term that
penalizes the total number of beacons with a weight
$\lambda$
\begin{equation}
  \label{eq:regdef}
  R(\{\ti_l\}) = \lambda \sum_l \ti_l^1.
\end{equation}

This setting simulates an environment that is complex enough to not
admit closed-form solutions for the inference function or the beacon
distribution.  Of course, there may be other phenomena in certain
applications, such as leakage across channels, multi-path
interference, etc., that are not modeled here. However, these too can
be incorporated in our framework as long as they can be modeled with an
appropriate environment function $\ee$.
\begin{figure*}[!t]
    \centering
    \subfigure[\scriptsize Learned: Annealed Reg. ($0.0497$)]{\rotatebox{90}{\scriptsize \bf ~~~~~~~~~~~~~Map 1}\includegraphics[width=0.245\textwidth]{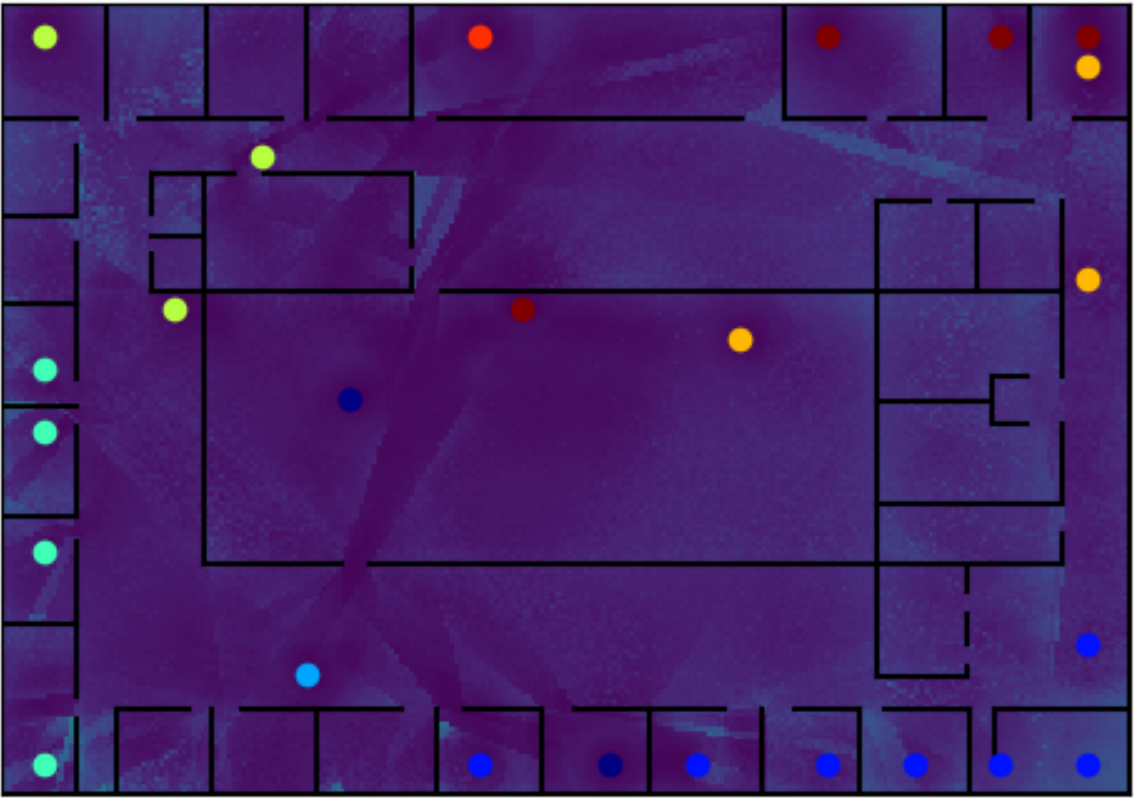}}\hfil
    \subfigure[\scriptsize Learned: High Fixed Reg. ($0.0633$)]{\includegraphics[width=0.245\textwidth]{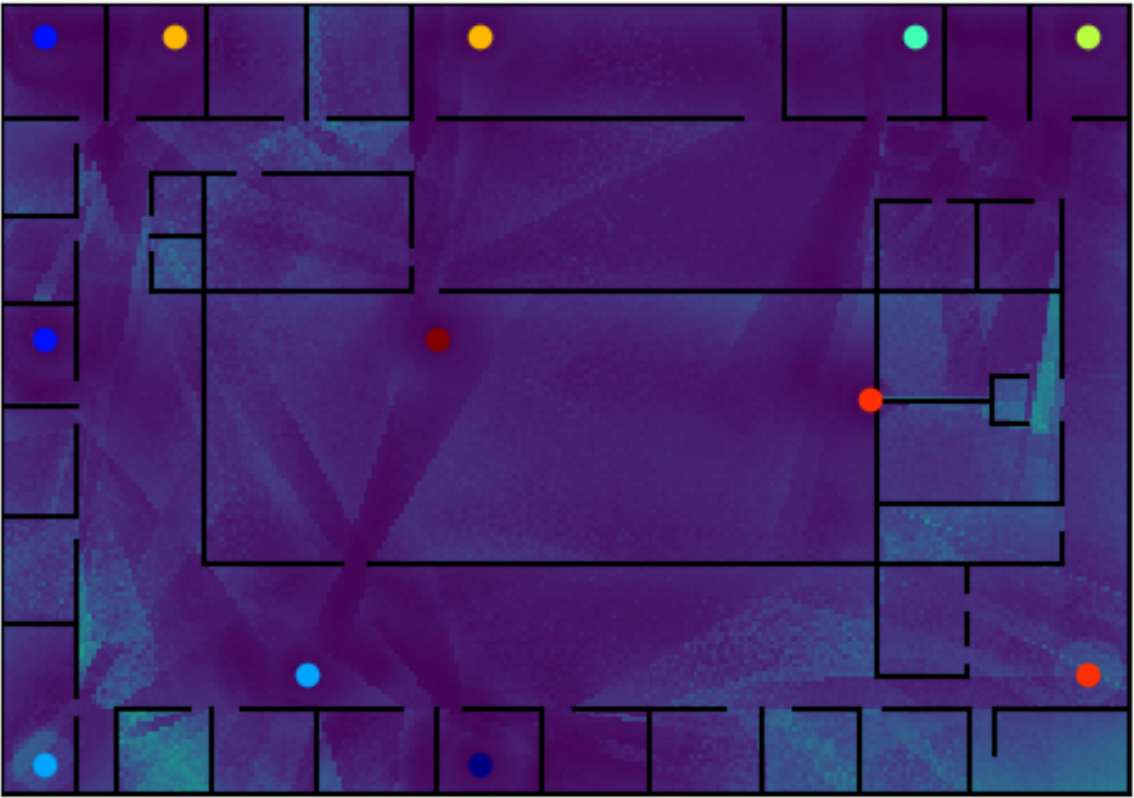}}\hfil
    \subfigure[\scriptsize Learned: Low Fixed Reg. ($0.0567$)]{\includegraphics[width=0.245\textwidth]{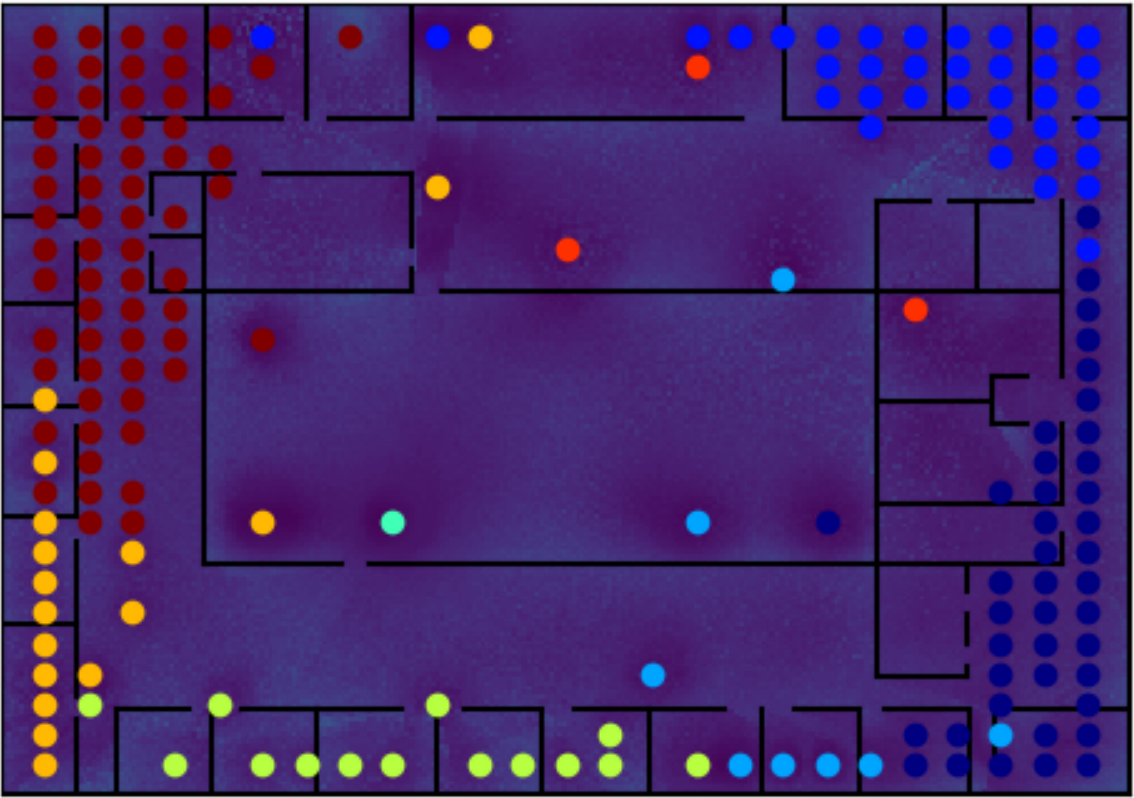}}\hfil
    \subfigure[\scriptsize Handcrafted A ($0.0716$)]{\includegraphics[width=0.245\textwidth]{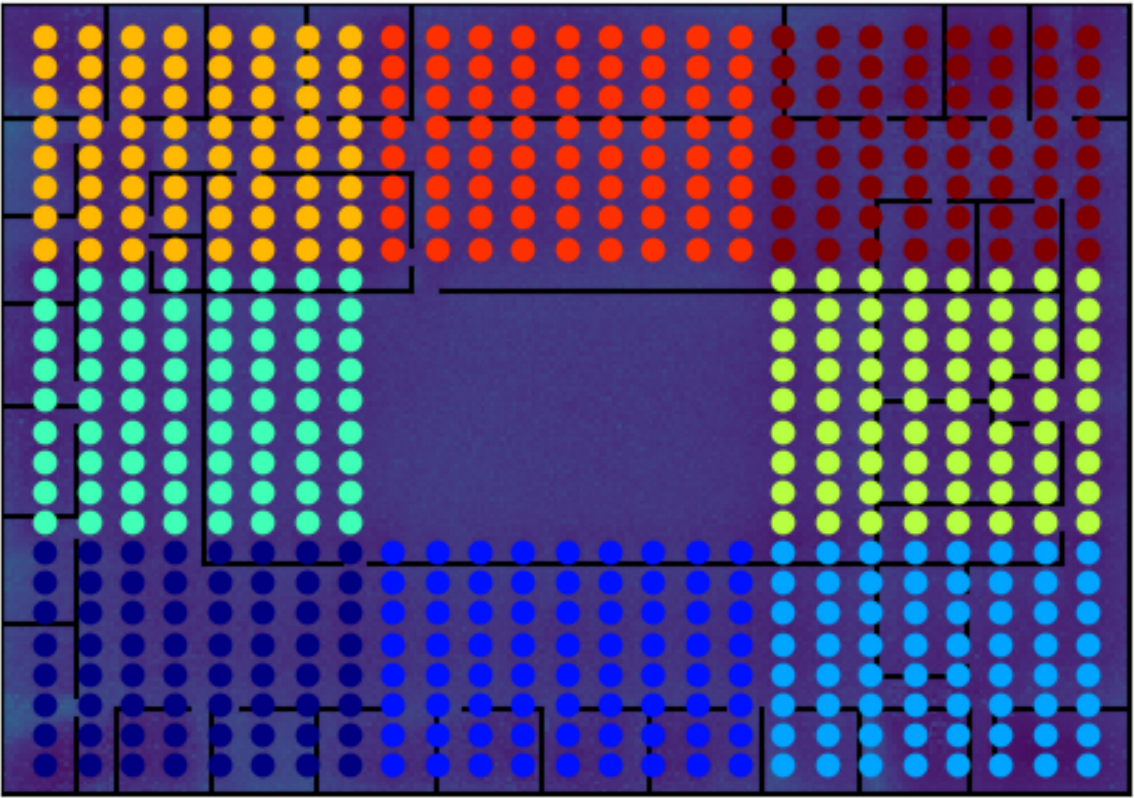}}\vspace{-0.75em}\\

    \subfigure[\scriptsize Learned: Annealed Reg. ($0.0473$)]{\rotatebox{90}{\scriptsize \bf ~~~~~~~~~~~~~Map 2}\includegraphics[width=0.245\textwidth]{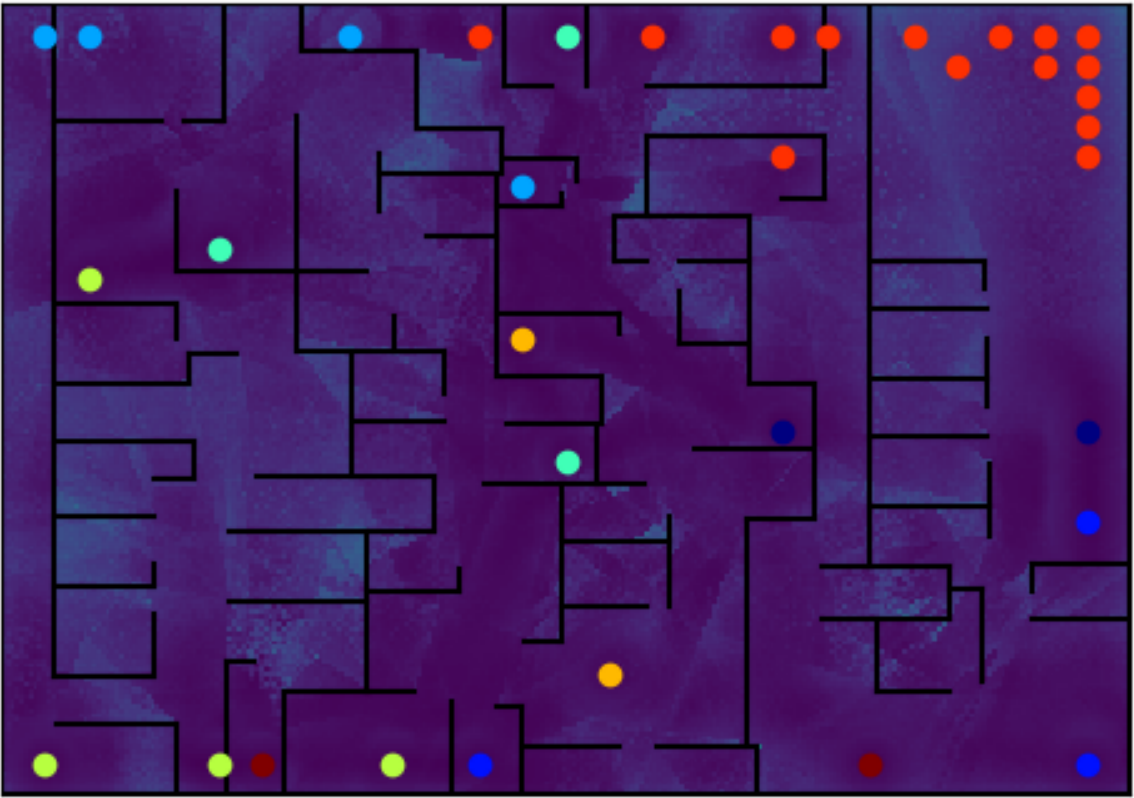}}\hfil
    \subfigure[\scriptsize Learned: High Fixed Reg.  ($0.0688$)]{\includegraphics[width=0.245\textwidth]{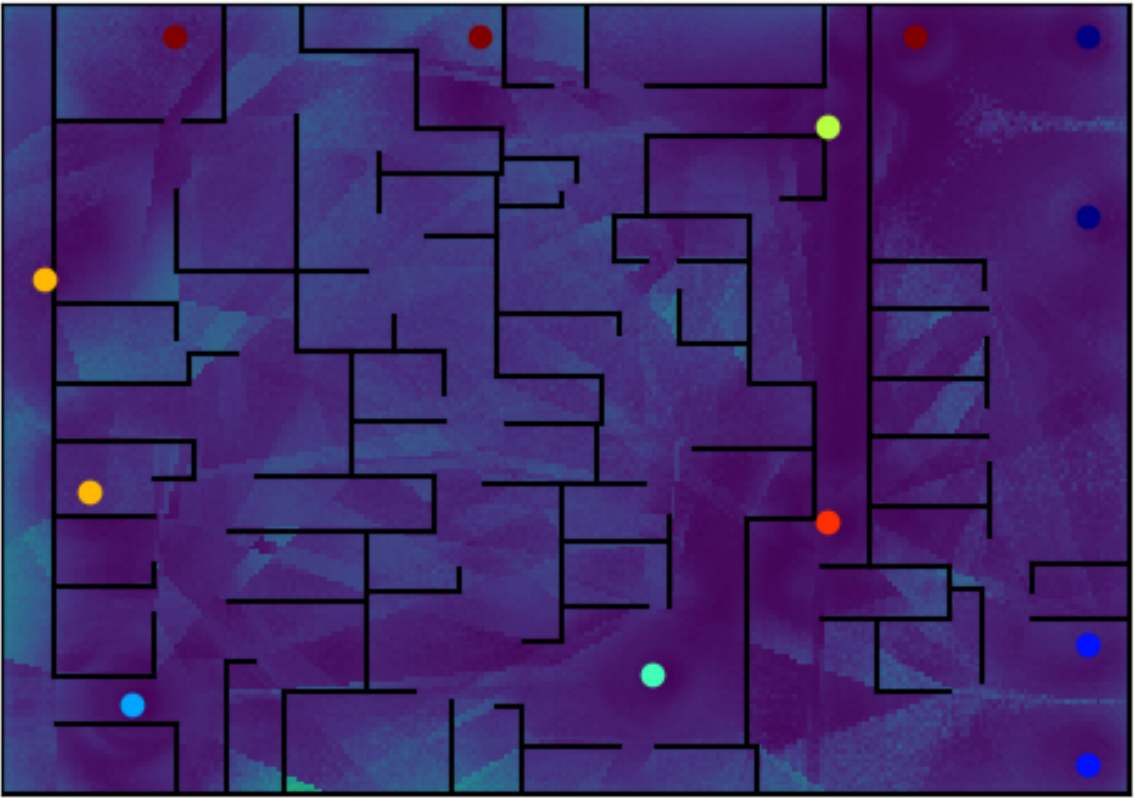}}\hfil
    \subfigure[\scriptsize Learned: Low Fixed Reg. ($0.04907$)]{\includegraphics[width=0.245\textwidth]{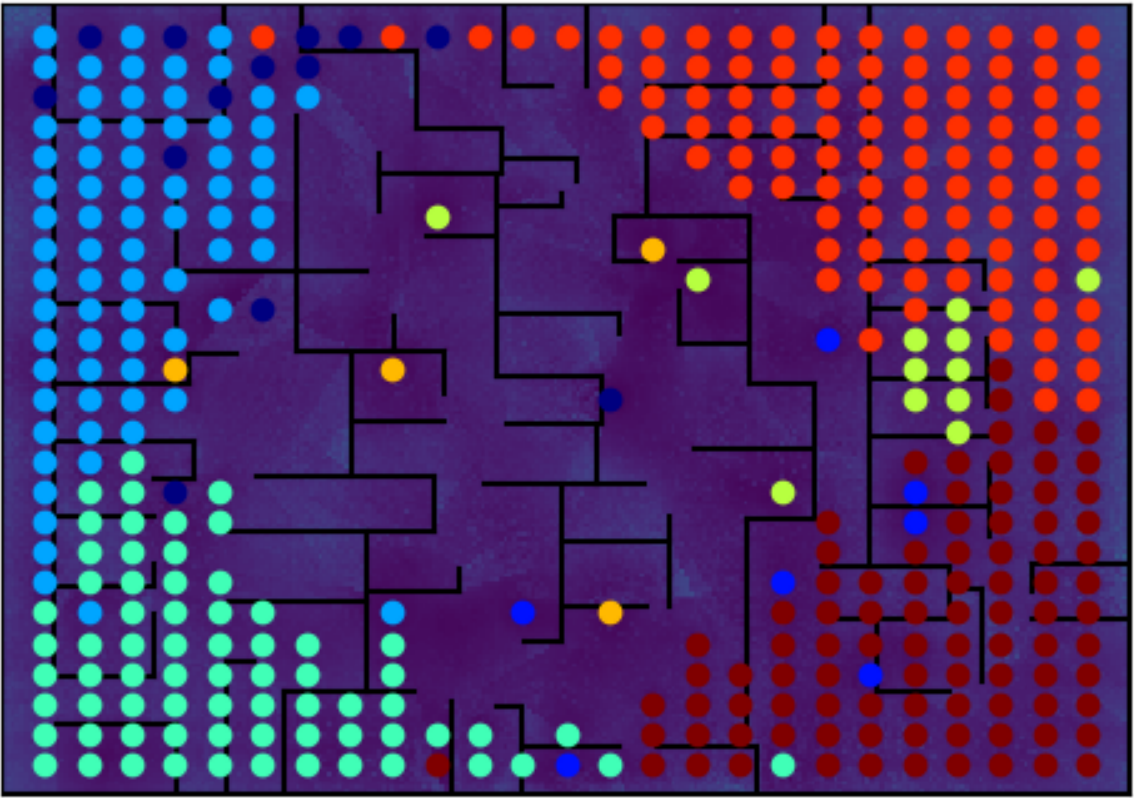}}\hfil
    \subfigure[\scriptsize Handcrafted B ($0.06242$)]{\includegraphics[width=0.245\textwidth]{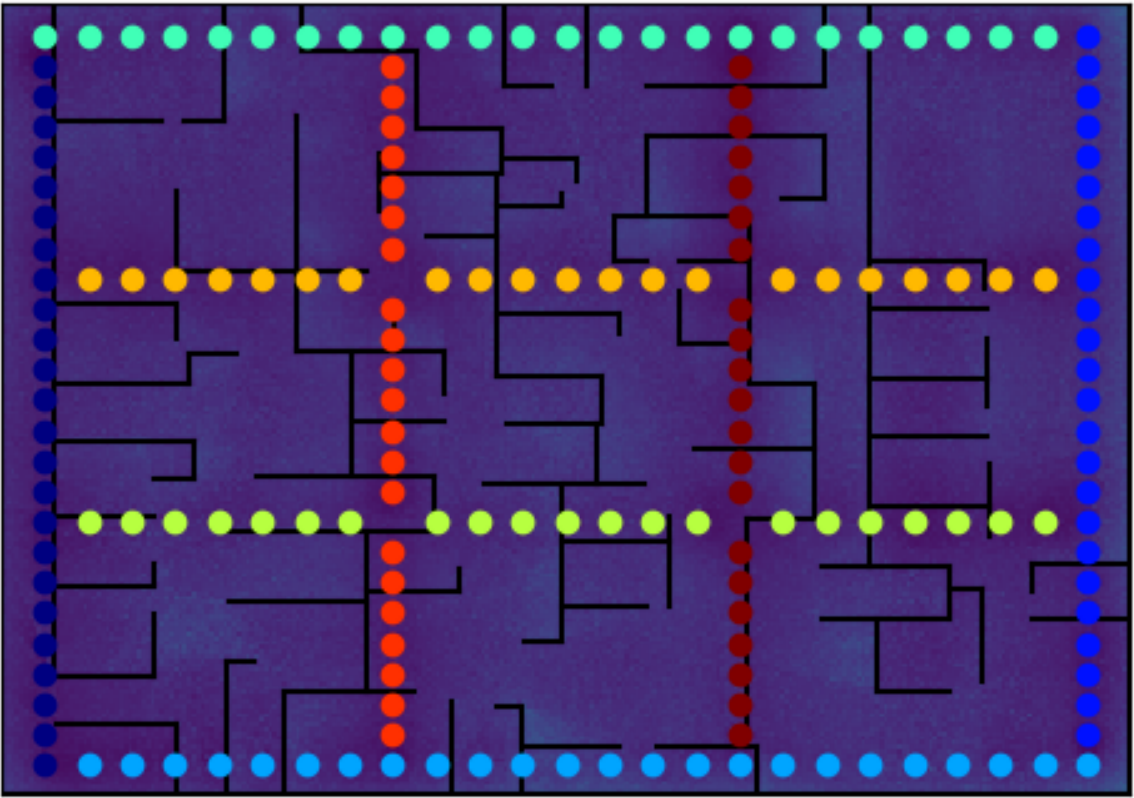}}\vspace{-0.75em}\\

    \subfigure[\scriptsize Learned: Annealed Reg. ($0.0493$)]{\rotatebox{90}{\scriptsize \bf ~~~~~~~~~~~~~Map 3}\includegraphics[width=0.245\textwidth]{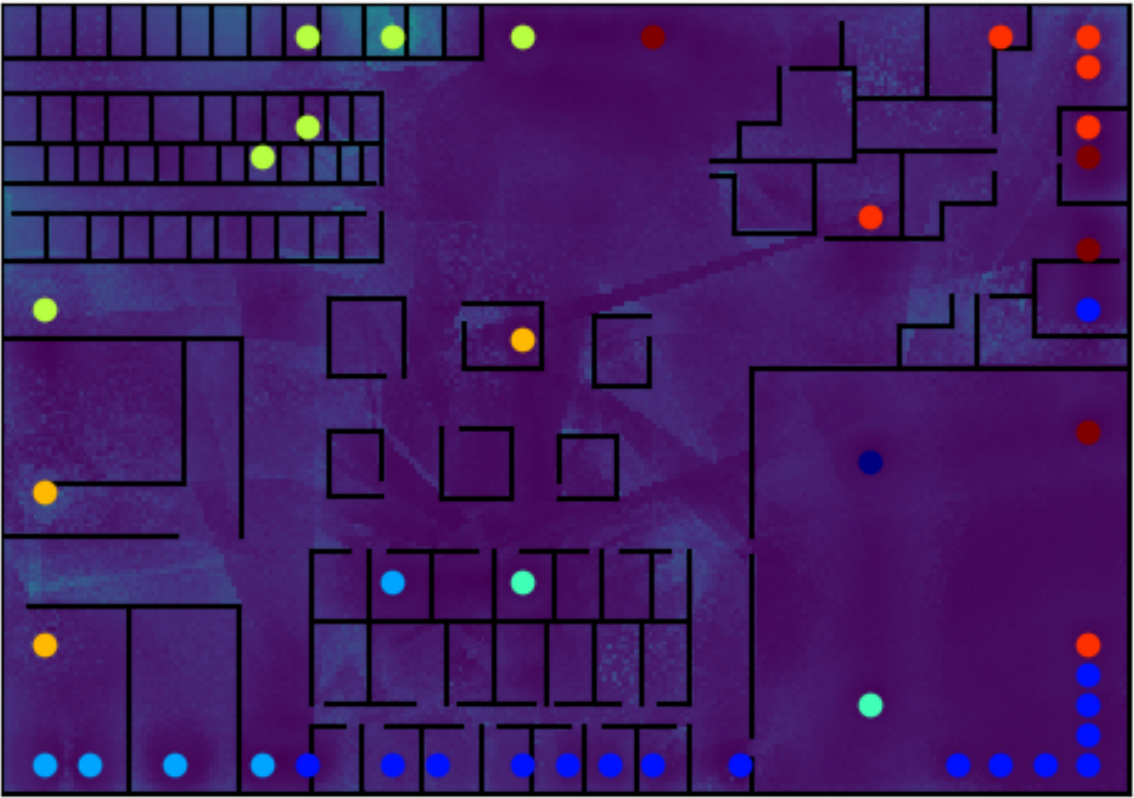}}\hfil
    \subfigure[\scriptsize Learned: High Fixed Reg.  ($0.0680$)]{\includegraphics[width=0.245\textwidth]{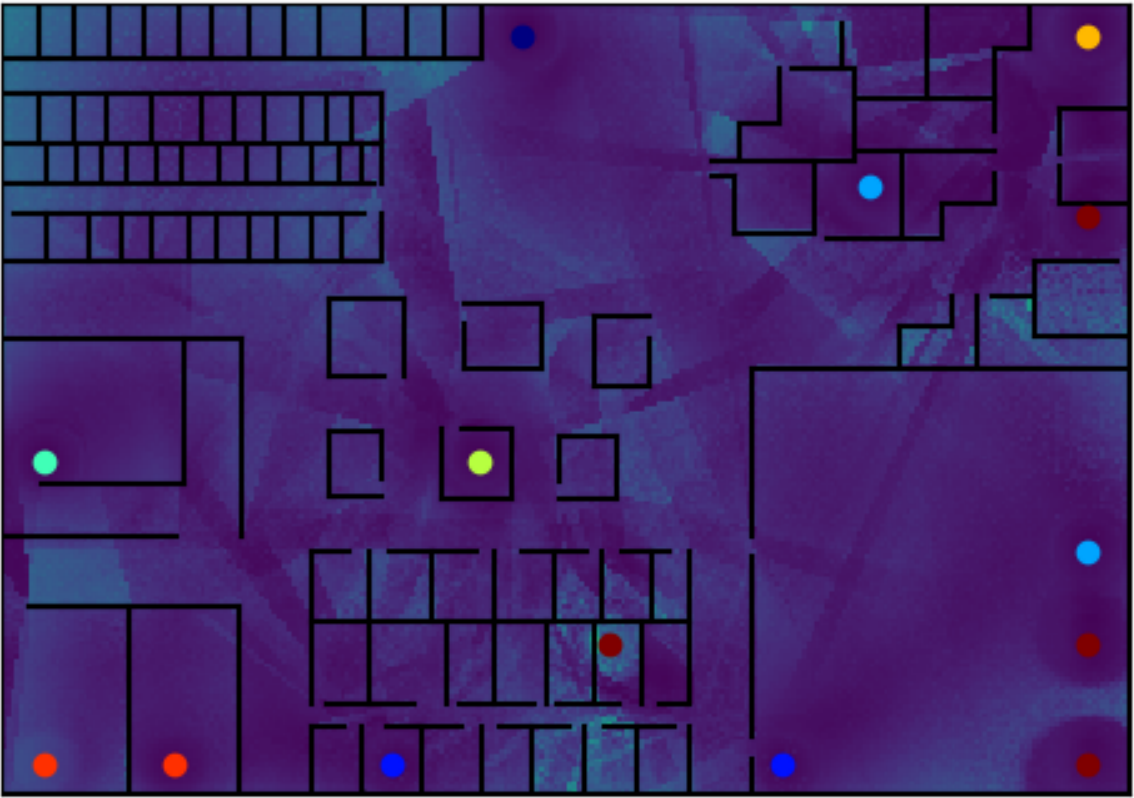}}\hfil
    \subfigure[\scriptsize Learned: Low Fixed Reg. ($0.0506$)]{\includegraphics[width=0.245\textwidth]{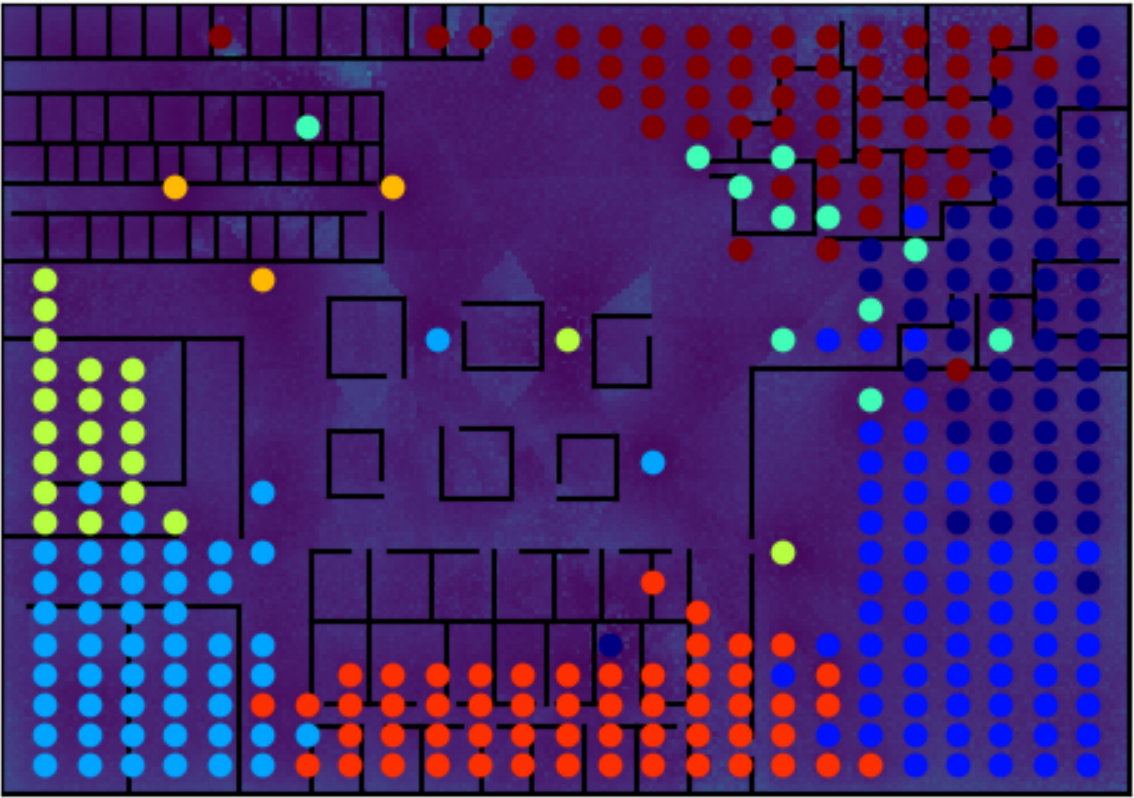}}\hfil
    \subfigure[\scriptsize Handcrafted B ($0.0649$)]{\includegraphics[width=0.245\textwidth]{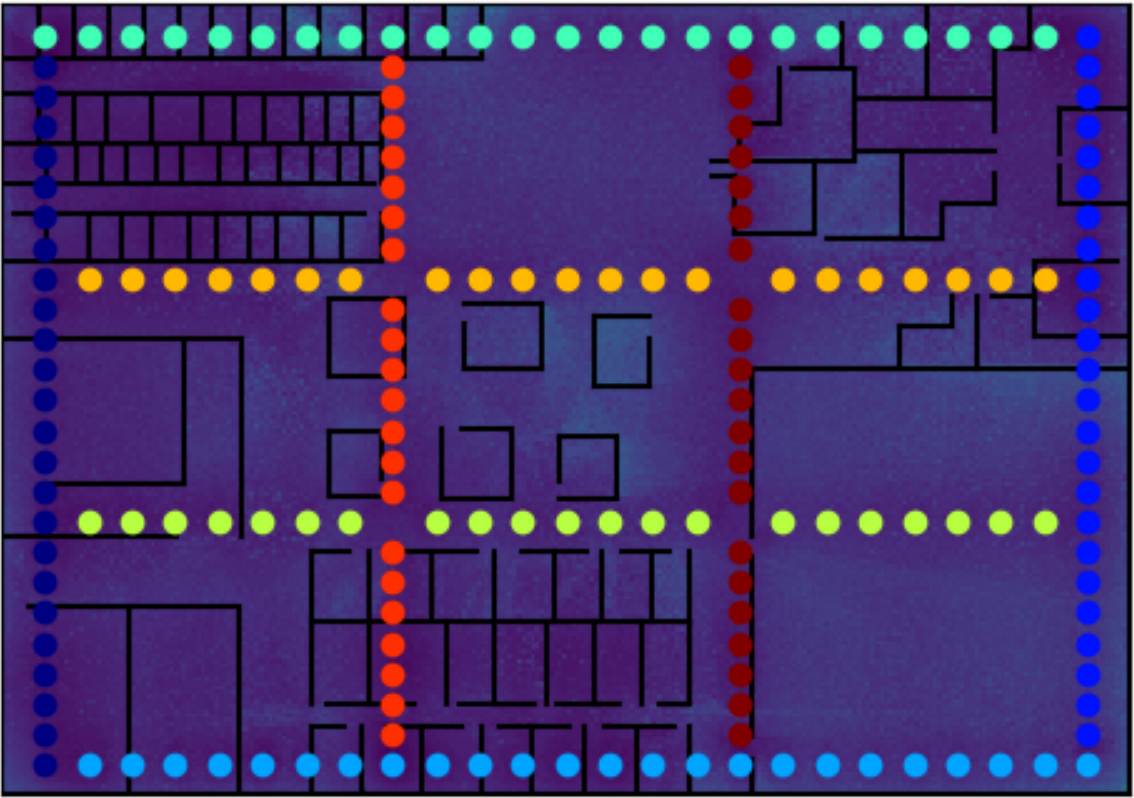}}

    \caption{Localization error maps for learned and handcrafted beacon distributions (with $8$ channels) for three environments. We show three learned distributions for each map---learned with annealed regularization, and two settings (high, low) of fixed regularization---as well as the best performing handcrafted placement. The overall RMSE is indicated below each error map (see Fig.~\ref{fig:motivation} for color mapping for errors).}
     \label{fig:map1-error}
\end{figure*}

\section{Results} \label{sec:results}

In this section, we evaluate our method through a series of simulation-based experiments on three different environment maps.

As discussed in Section~\ref{sec:related}, existing methods are not well-suited to beacon placement for localization. Consequently, we compare our method to to several hand-designed beacon allocation strategies. We first show that our inference network is effective at localization given the baseline placements by comparing to a standard nearest-neighbors method. We then analyze the performance when learning the beacon allocation and inference network jointly, demonstrating that our method learns a distribution and inference strategy that enable high localization accuracy for all three environments. We end by analyzing the effects of different degrees of regularization and different numbers of available channels, as well as the variation in the learned beacon distributions based on parameters of the environment function $\ee$.

\subsection{Setup}

We conduct our experiments on three manually drawn environment maps,
which correspond to floor plans (of size $1\times 0.7$ map units)
with walls that serve as obstructions. For each map, we arrange $L=625$
possible beacon locations in a $25 \times 25$ evenly spaced grid. We
consider configurations with values of $C=4,8,$ and $16$ RF-channels.

Our experiments use the environment model defined in Eqn.~\ref{eq:plv}
with $P_0 = 6.25\times 10^{-4}$, $\zeta=2.0$ (where locations are in
map units), and $\beta = e^{-1.0}$, with sensor noise variance
$\sigma_z^2=10^{-4}$. The sensor measurements are saturated at a
threshold $\tau=1.0$.

While training the beacons, we use parameters $\alpha_0 = 1$ and $\gamma = 1.25\times10^{-9}$ for the quadratic temperature scheme. These
values were chosen empirically so that the beacon selection vectors
$\tilde{I}_l$ converge at the same pace as it takes the inference
network to learn (as observed while training on a fixed beacon
distribution).  After $900$k iterations, we switch the softmax
to an ``arg-max'', effectively setting $\alpha$ to infinity and fixing
the beacon placement, and then continue training the inference network.

The inference function $f(\cdot)$ is parameterized
as a $13$-layer feed-forward neural network. Our architecture consists
of $6$ blocks of $2$ fully-connected layers. All hidden layers contain
$1024$ units and are followed by ReLU activation. Each block is
followed by a max pooling operation applied on disjoint sets of $4$
units. After the last block, there is a final output layer with $2$
units that predicts the $(x,y)$ location coordinates.

During training, locations are randomly sampled and fed through our
environment model to the inference network in batches of $1000$. All
networks are trained by minimizing the loss defined in
Eqn.~\ref{eq:lossdef} for $1000$k iterations using SGD with a
learning rate of $0.01$ and momentum $0.9$, followed by an additional $100$k iterations with a learning rate $0.001$. We also use
batch-normalization in all hidden layers.

\subsection{Experiments}

We evaluate the effectiveness of our approach for joint optimization of beacon placement and localization through a series of experiments. Localization performance can vary significantly within an environment. Consequently, for each beacon allocation and inference strategy, we report both average as well as worst-case performance over a dense set of locations, with multiple samples (corresponding to different random noise and interference phases) per location. We measure performance in terms of the root mean squared error (RMSE)---between estimated and true coordinates---over all samples at all locations, as well as over the worst sample at each location, which we refer to as \textit{worst-case RMSE}.  We also measure the frequency with which large errors occur---defined  with respect to different thresholds ($0.1, 0.2, 0.5$)---and report these as \textit{failure rates}.

\begin{table*}[!t]
    \centering
    \setlength{\tabcolsep}{4pt}
    \caption{Performance Comparison for Different Inference and Placement Strategies}\label{tab:overall}
    \begin{tabularx}{1.0\linewidth}{c l l c c c c c c}
        \toprule
        Map & Allocation & Inference & Beacons & RMSE & RMSE (Worst-case) & Failure Rate ($0.10$) & Failure Rate ($0.20$) & Failure Rate ($0.50$)\\
        \toprule
        \multirow{7}{*}{1} & Handcrafted A & kNN & $544$ & $0.0817$ & $0.1793$ & $19.1946~\%$ & $1.4662~\%$ & $0.0096~\%$\\
        & Handcrafted B & kNN & $180$ & $0.0998$ & $0.2384$ & $26.1523~\%$ & $4.7400~\%$ & $0.0600~\%$\\\cline{2-9}
        & Handcrafted A & Network & $544$ & $0.0716$ & $0.1537$ & $13.5691~\%$ & $0.6706~\%$ & $\mathbf{0.0045~\%}$\\
        & Handcrafted B & Network & $180$ & $0.0811$ & $0.1940$ & $17.1529~\%$ & $2.4632~\%$ & $0.0192~\%$\\\cline{2-9}
        & Learned (low fixed reg.) & Network & $183$ & $0.0567$ & $0.1336$ & $\hphantom{1}6.4917~\%$ & $0.4866~\%$ & $0.0053~\%$\\
        & Learned (high fixed reg.) & Network & $\hphantom{1}12$ & $0.0633$ & $0.1511$ & $\hphantom{1}8.0217~\%$ & $1.3686~\%$ & $0.0827~\%$\\
        & Learned (annealed reg.) & Network & $\hphantom{1}25$ & $\mathbf{0.0497}$ & $\mathbf{0.1169}$ & $\hphantom{1}\mathbf{4.7705}~\%$ & $\mathbf{0.3706}~\%$ & $0.0125~\%$\\
        \toprule
        \multirow{7}{*}{2} & Handcrafted A & kNN & $544$ & $0.0806$ & $0.1708$ & $18.7127~\%$ & $1.3098~\%$ & $0.0023~\%$\\
        & Handcrafted B & kNN & $180$ & $0.0839$ & $0.2040$ & $18.3838~\%$ & $2.4336~\%$ & $0.0306~\%$\\\cline{2-9}
        & Handcrafted A & Network & $544$ & $0.0653$ & $0.1331$ & $10.4884~\%$ & $\mathbf{0.2718~\%}$ & $\mathbf{0.0001~\%}$\\
        & Handcrafted B & Network & $180$ & $0.0624$ & $0.1479$ & $\hphantom{1}9.1473~\%$ & $0.6843~\%$ & $0.0013~\%$\\\cline{2-9}
        & Learned (low fixed reg.) & Network & $371$ & $0.0491$ & $0.1154$ & $\hphantom{1}\mathbf{4.3015~\%}$ & $0.2931~\%$ & $0.0016~\%$\\
        & Learned (high fixed reg.)& Network & $\hphantom{1}13$ & $0.0688$ & $0.1450$ & $11.9567~\%$ & $1.4807~\%$ & $0.0040~\%$\\
        & Learned (annealed reg.)& Network & $\hphantom{1}35$ & $\mathbf{0.0473}$ & $\mathbf{0.1142}$ & $\hphantom{1}4.6226~\%$ & $0.4182~\%$ & $0.0091~\%$\\
        \toprule
        \multirow{7}{*}{3} & Handcrafted A & kNN & $544$ & $0.0814$ & $0.1717$ & $18.7939~\%$ & $1.6503~\%$ & $0.0027~\%$\\
        & Handcrafted B & kNN & $180$ & $0.0840$ & $0.2127$ & $16.6865~\%$ & $2.6309~\%$ & $0.0978~\%$\\\cline{2-9}
        & Handcrafted A & Network & $544$ & $0.0670$ & $0.1432$ & $11.1586~\%$ & $0.6904~\%$ & $\mathbf{0.0009~\%}$\\
        & Handcrafted B & Network & $180$ & $0.0649$ & $0.1581$ & $\hphantom{1}9.7563~\%$ & $1.0512~\%$ & $0.0079~\%$\\\cline{2-9}
        & Learned (low fixed reg.) & Network & $325$ & $0.0506$ & $0.1200$ & $\hphantom{1}\mathbf{4.2496~\%}$ & $\mathbf{0.3482~\%}$ & $0.0056~\%$\\
        & Learned (high fixed reg.)& Network & $\hphantom{1}14$ & $0.0680$ & $0.1495$ & $11.6651~\%$ & $1.2166~\%$ & $0.0454~\%$\\
        & Learned (annealed reg.)& Network & $\hphantom{1}43$ & $\mathbf{0.0493}$ & $\mathbf{0.1158}$ & $\hphantom{1}4.7211~\%$ & $0.6529~\%$ & $0.0045~\%$\\
        \bottomrule
    \end{tabularx}
\end{table*}

Table~\ref{tab:overall} reports these metrics on the three environment maps for different versions our approach that vary the regularization settings. We have also experimented with a number of manually handcrafted distribution strategies for these maps and report the performance of the two strategies that worked best in Table~\ref{tab:overall}. For the handcrafted settings, we report the result of training a neural network for inference, as well as of $k$-nearest neighbors (kNN)-based inference (we try $k\in \{1,5,10,20\}$ and pick the best). These latter experiments show that the network-based inference performs well (better than the kNN baseline) and, therefore, that our architecture is reasonable for the task. Moreover, the results reveal that jointly optimizing beacon allocation and inference provides accuracies that exceed the  handcrafted baselines, yielding different distribution strategies with different numbers of beacons. Figure~\ref{fig:map1-error} visualizes the beacon placement and channel allocation along with the RMSE for both learned and handcrafted beacon allocations.

\begin{figure}[!t]
    \centering
    \includegraphics[width=\linewidth]{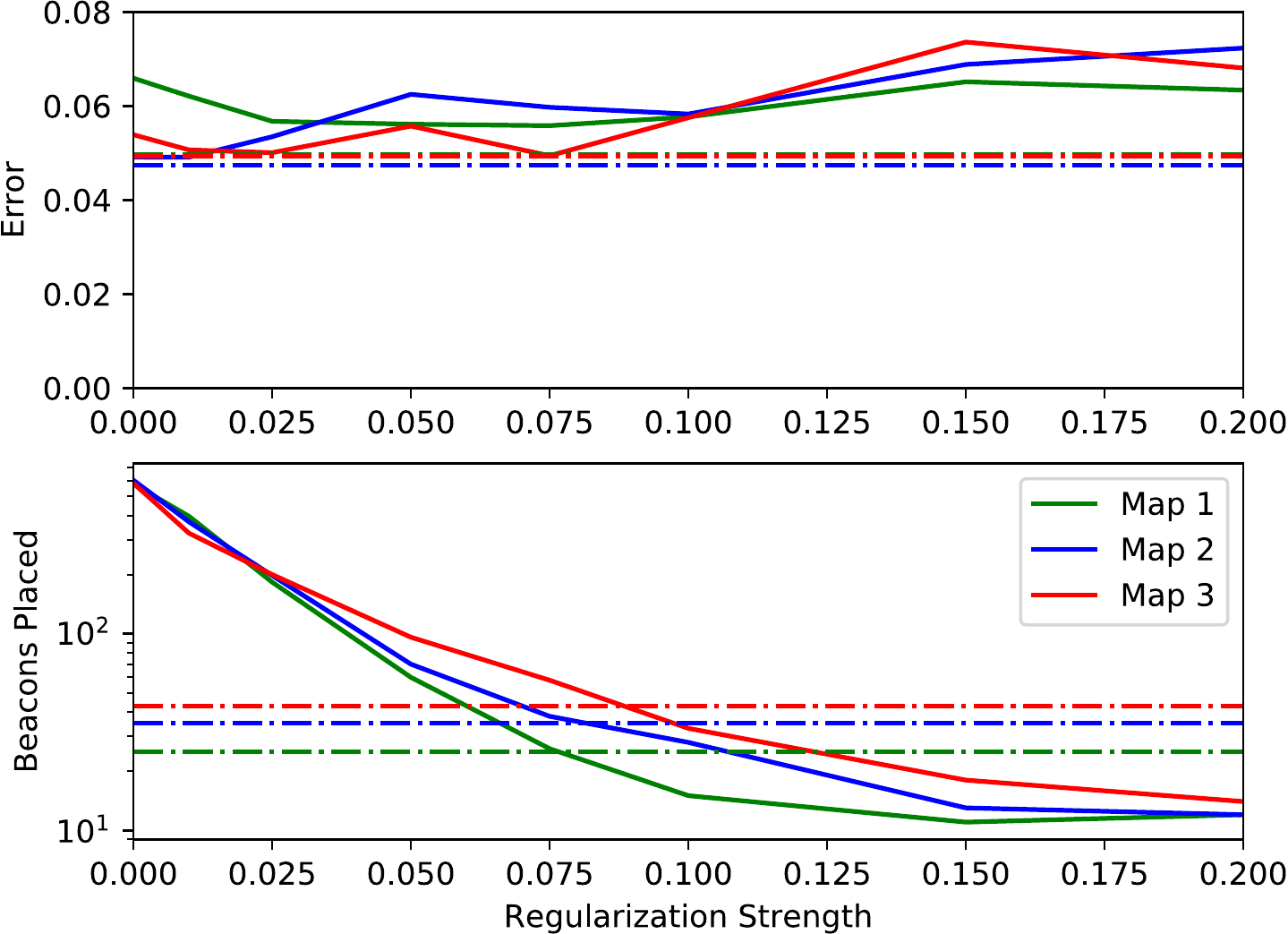}
    \caption{A plot showing the effects of regularization on mean error and the number of beacons placed. The dashed lines represent the mean error and beacons placed with annealed regularization on each map.} \label{fig:reg}
\end{figure}

Next, we report a more detailed evaluation of the regularization scheme defined in Eqn.~\ref{eq:regdef}, and therefore the ability of our method to allow for a trade-off between the number of beacons placed and accuracy. First, we use a constant value of $\lambda$, trying various values between $0.0$ and $0.2$. As Figure~\ref{fig:reg} shows, increased regularization leads to solutions with fewer beacons. On Map 2, we find that decreased regularization always leads to solutions with lower error. On Maps 1 and 3, however, unregularized beacon placement results in increased localization error. This suggests that regularization may also allow our model to escape bad local minima during training. Then, we experiment with an annealing scheme for $\lambda$ and find that it leads to a better performance-cost trade-off. We use a simple annealing schedule that decays $\lambda = 0.2$ by a constant factor $\eta = 0.25$ every $100$k iterations.

\begin{figure*}[!t]
    \centering
    {\scriptsize \bf Map 1}\\
    \subfigure[\scriptsize Iter \# $10$k]{\includegraphics[width=0.192\textwidth]{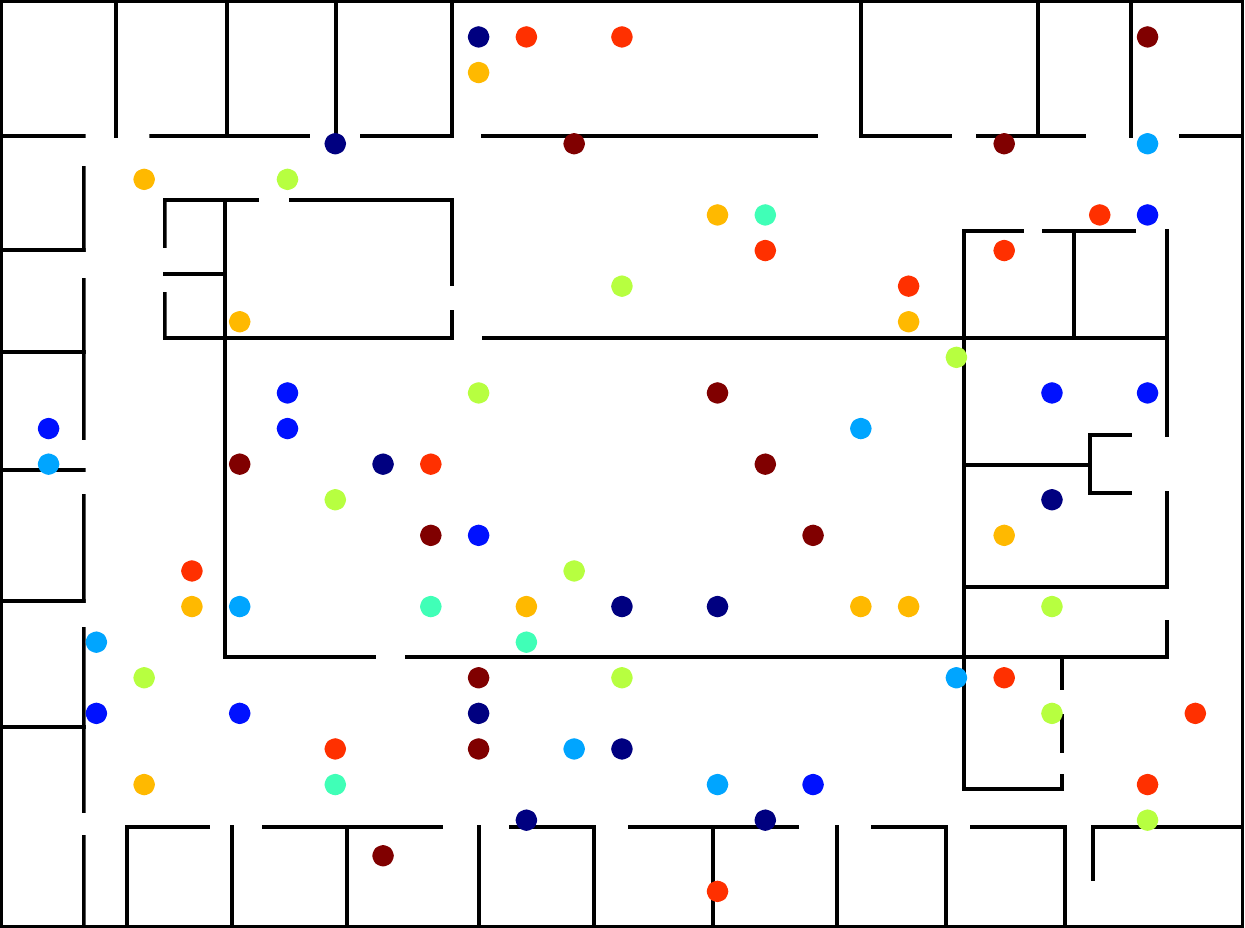}}\hfil
    \subfigure[\scriptsize Iter \# $20$k]{\includegraphics[width=0.192\textwidth]{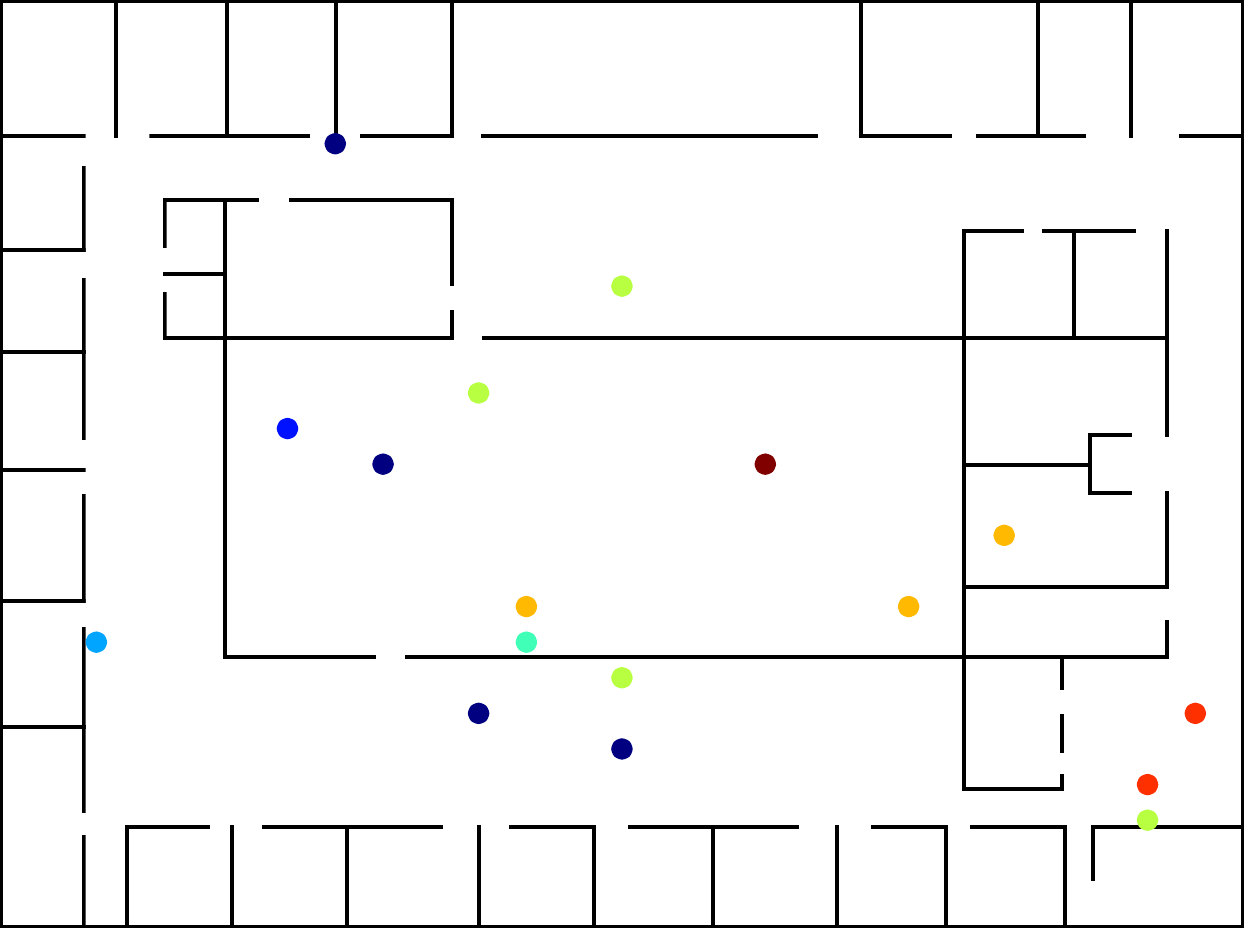}}\hfil
    \subfigure[\scriptsize Iter \# $40$k]{\includegraphics[width=0.192\textwidth]{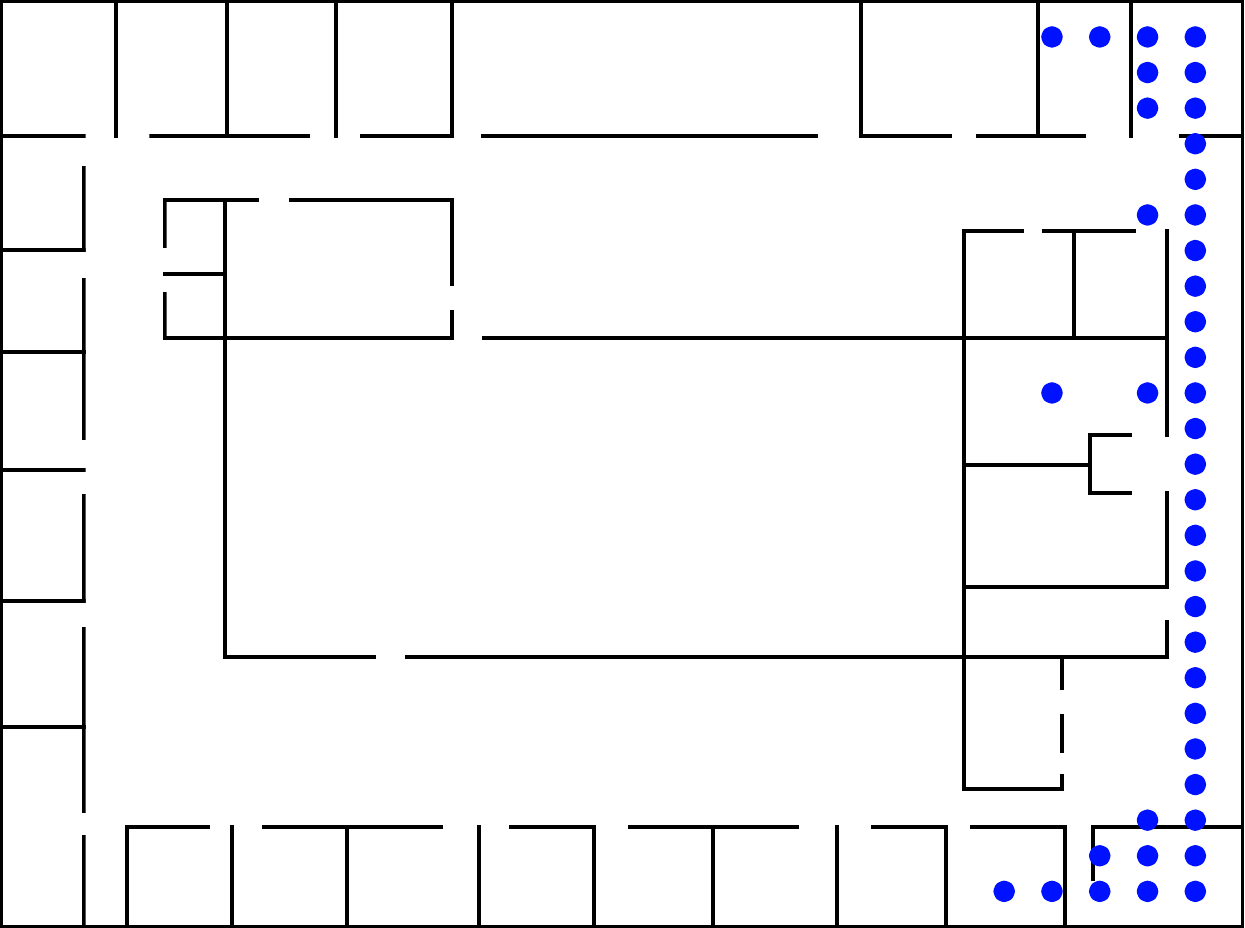}}\hfil
    \subfigure[\scriptsize Iter \# $70$k]{\includegraphics[width=0.192\textwidth]{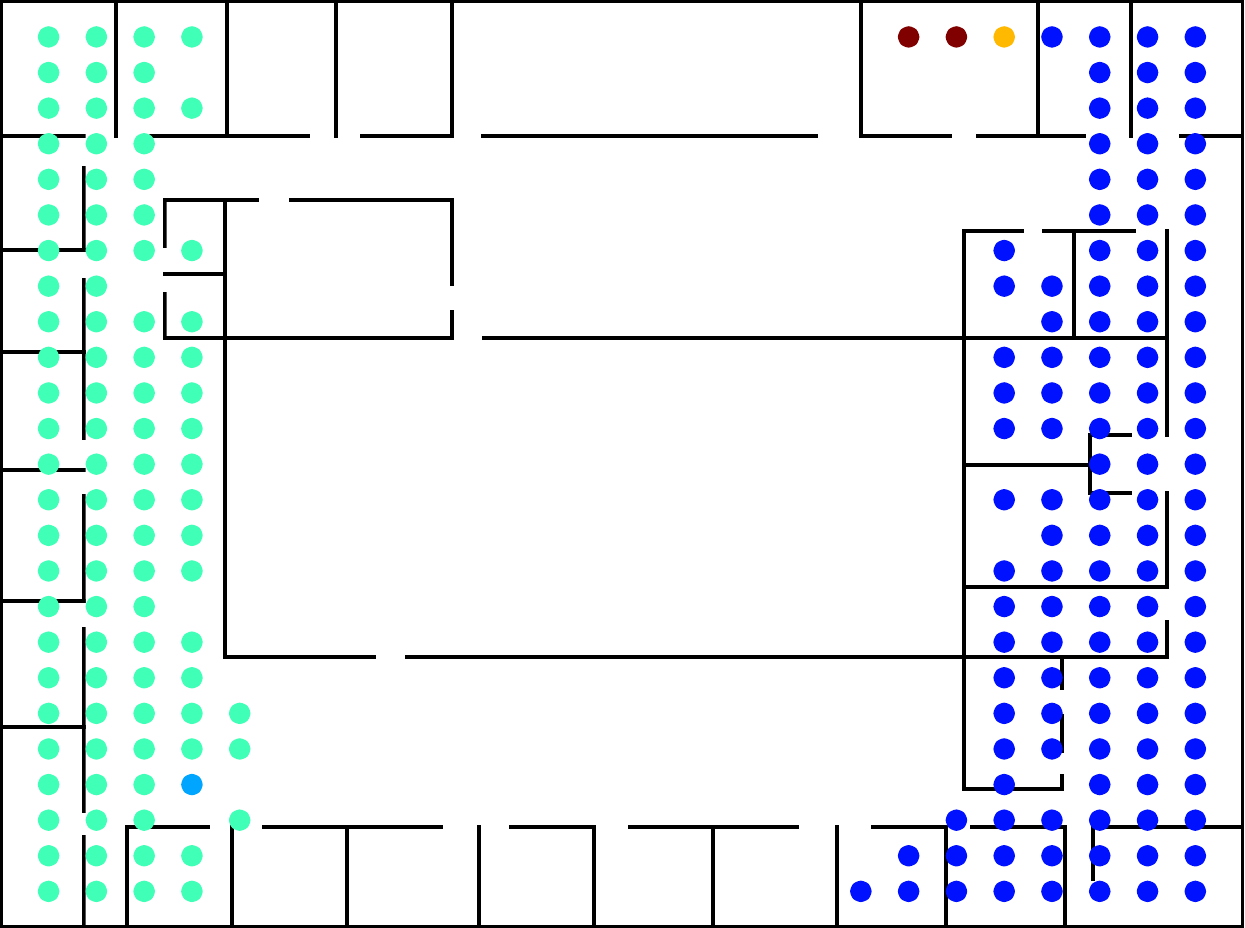}}\hfil
    \subfigure[\scriptsize Iter \# $110$k]{\includegraphics[width=0.192\textwidth]{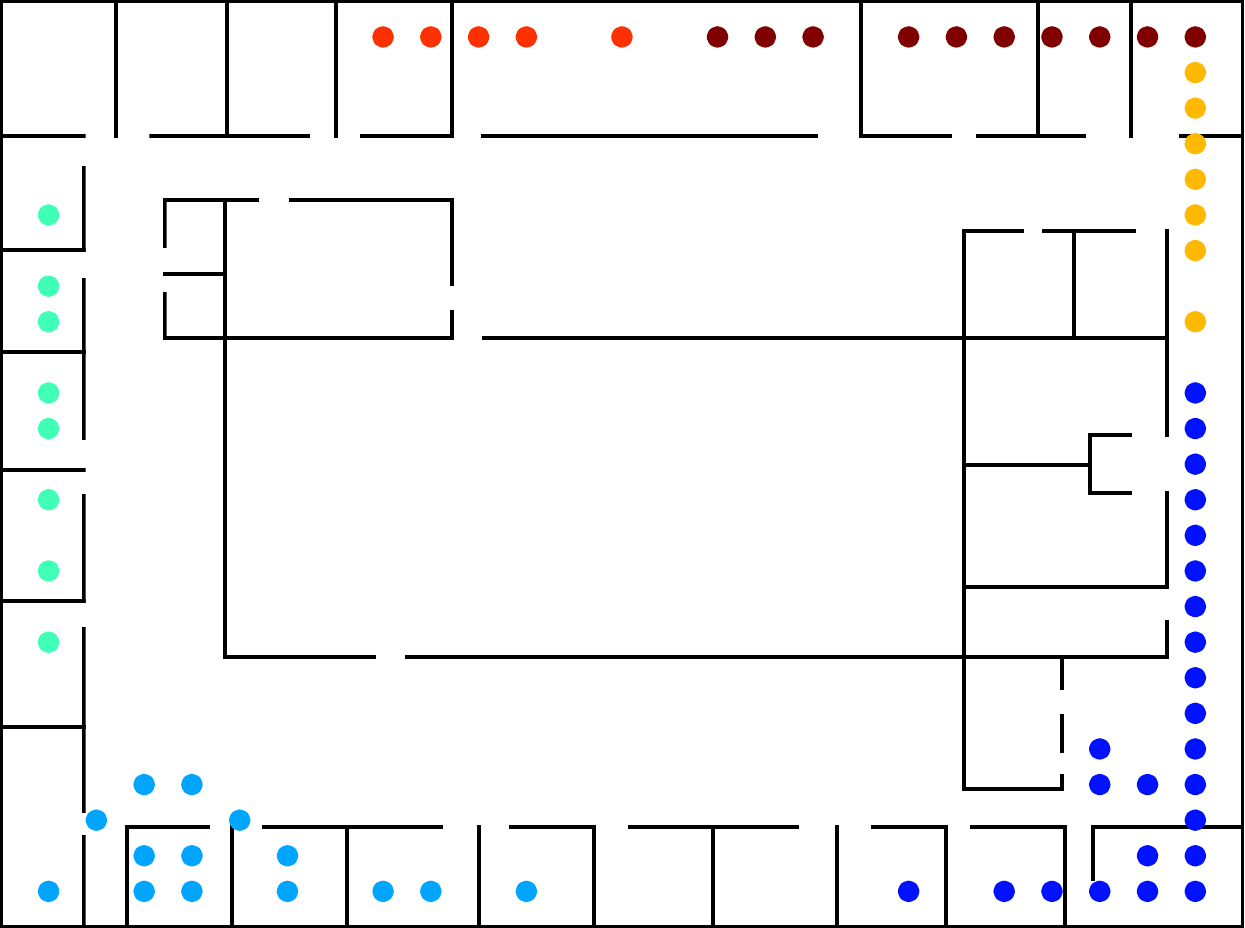}}\vspace{-0.65em}\\
    
    \subfigure[\scriptsize Iter \# $180$k]{\includegraphics[width=0.192\textwidth]{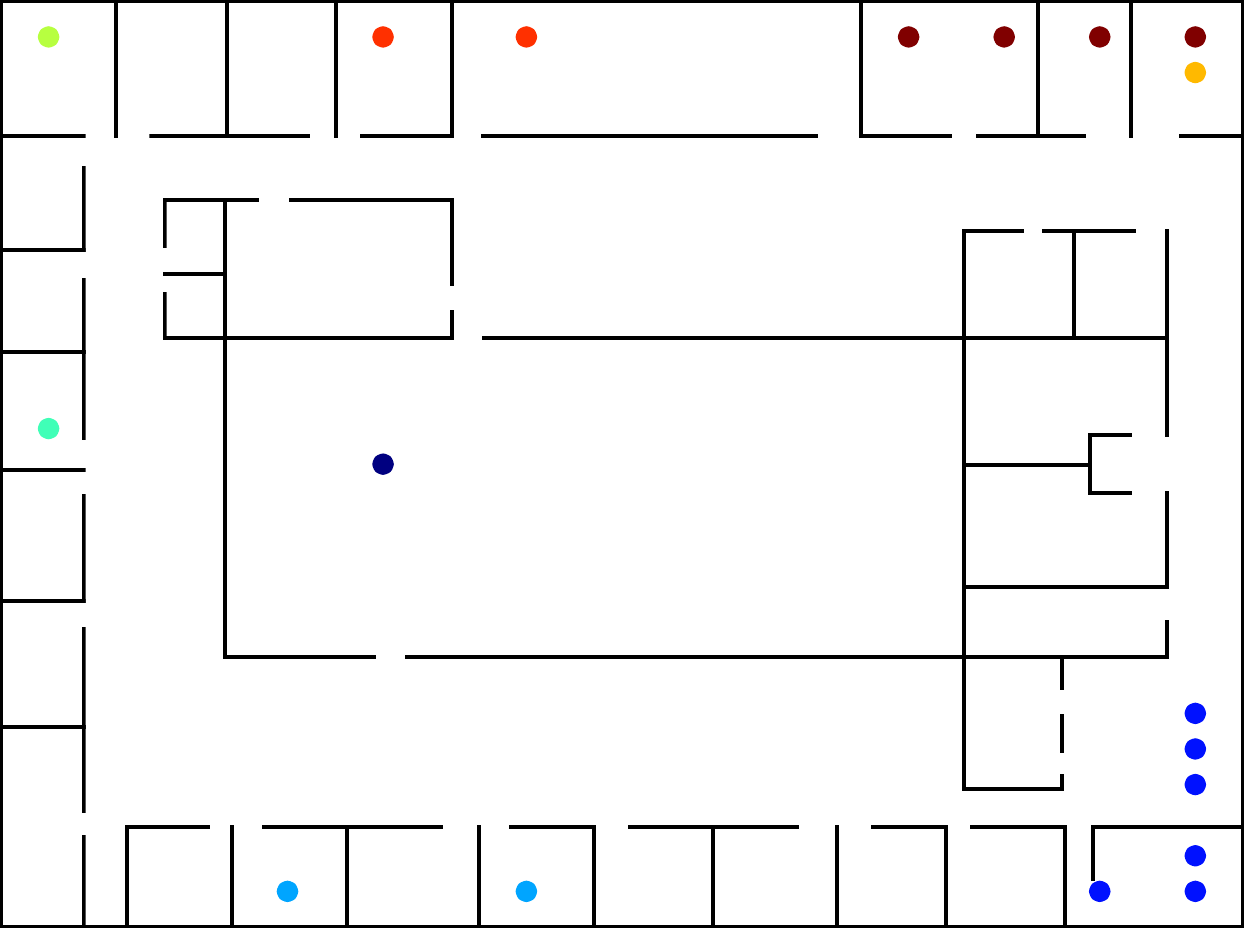}}\hfil
    \subfigure[\scriptsize Iter \# $250$k]{\includegraphics[width=0.192\textwidth]{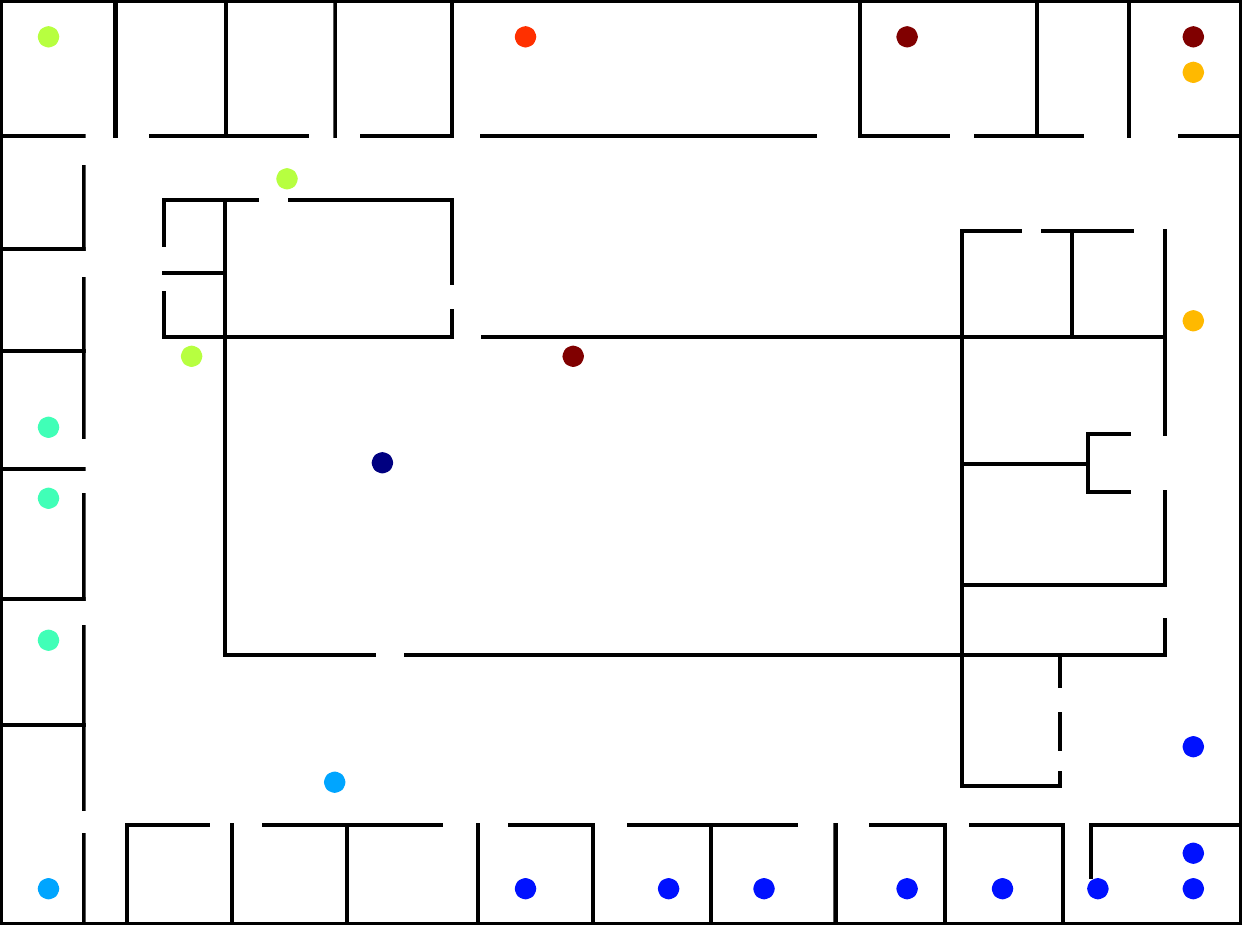}}\hfil
    \subfigure[\scriptsize Iter \# $320$k]{\includegraphics[width=0.192\textwidth]{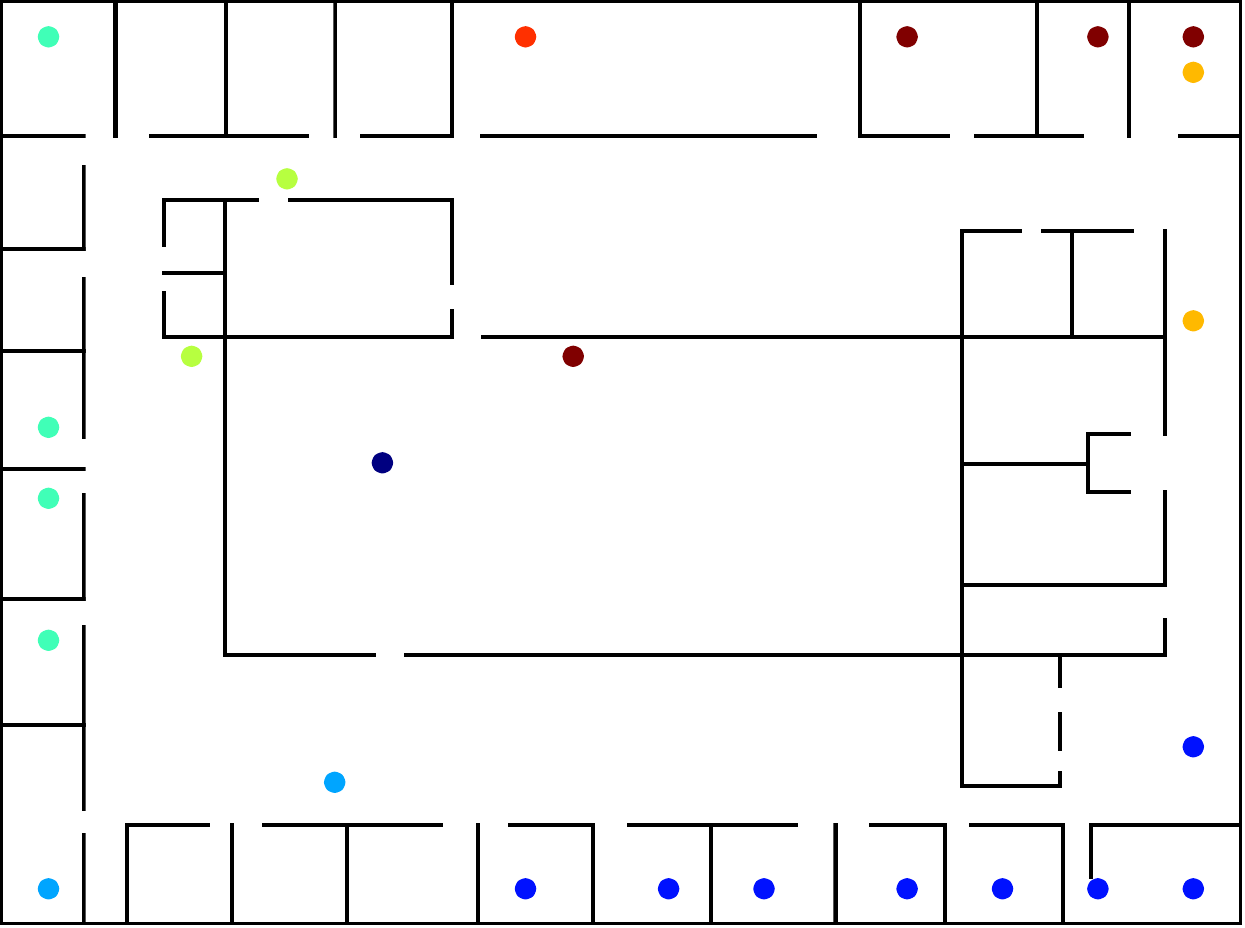}}\hfil
    \subfigure[\scriptsize Iter \# $500$k]{\includegraphics[width=0.192\textwidth]{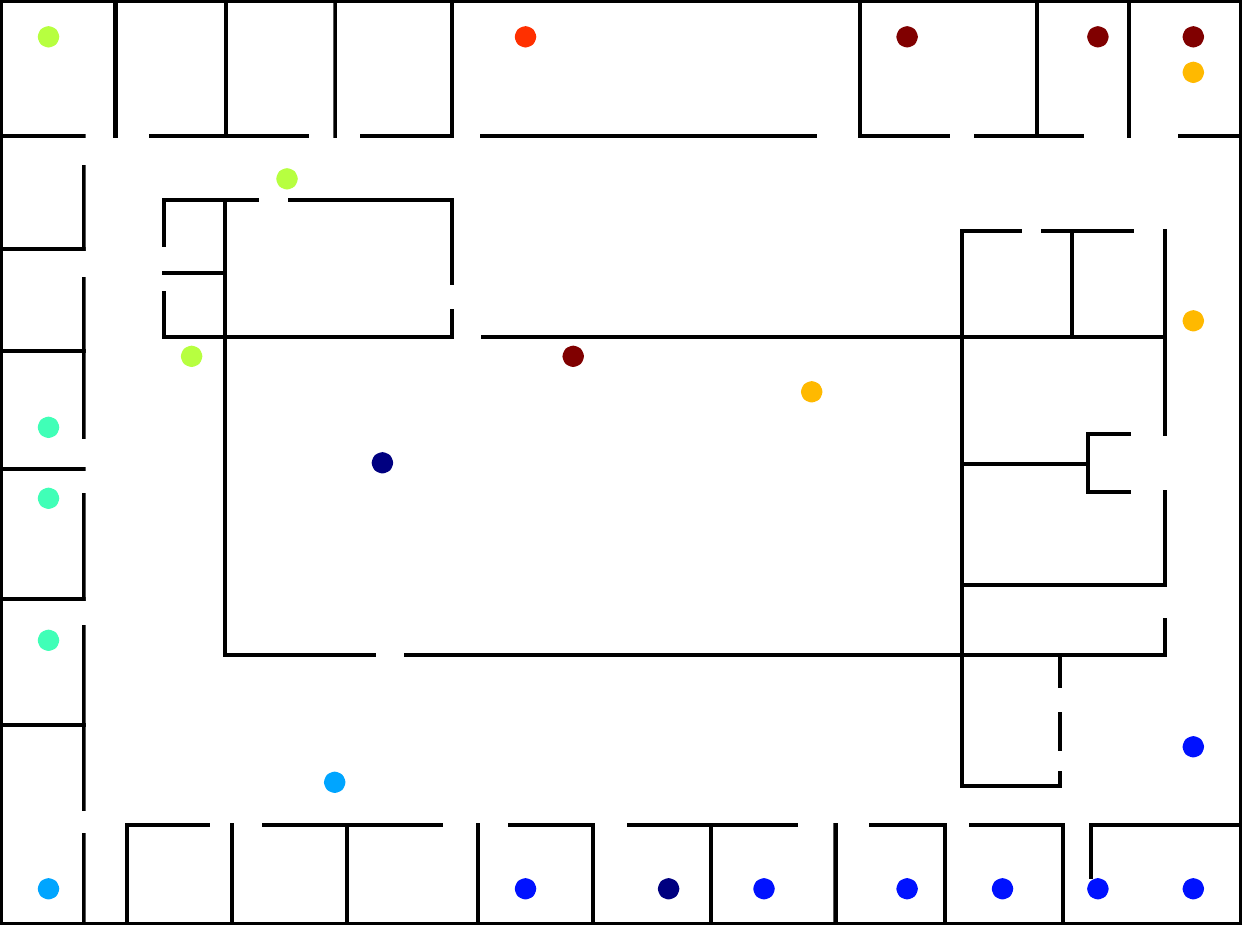}}\hfil
    \subfigure[\scriptsize Final]{\includegraphics[width=0.192\textwidth]{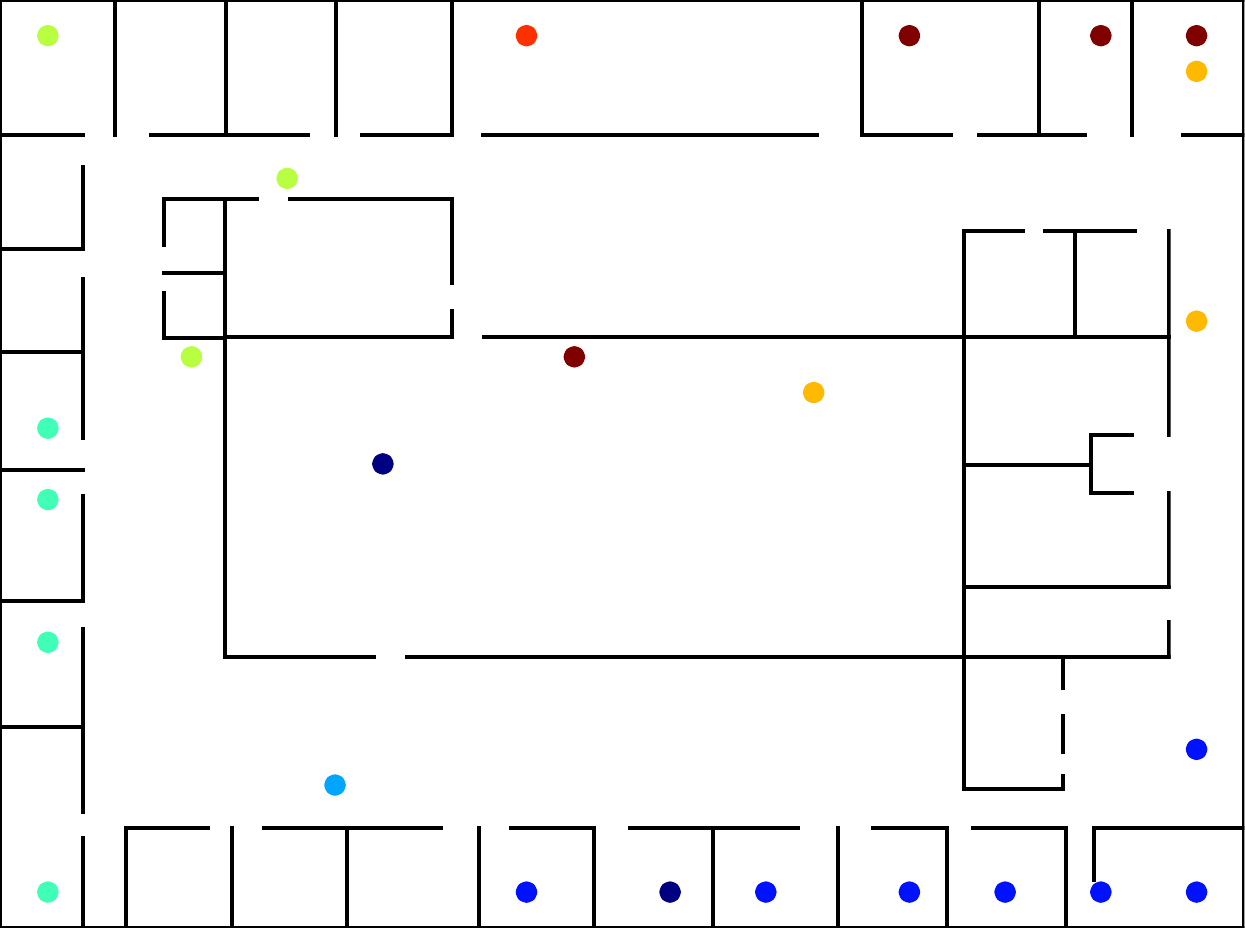}}\vspace{0.9em}\\

    {\scriptsize\bf Map 2}\\
    \subfigure[\scriptsize Iter \# $10$k]{\includegraphics[width=0.192\textwidth]{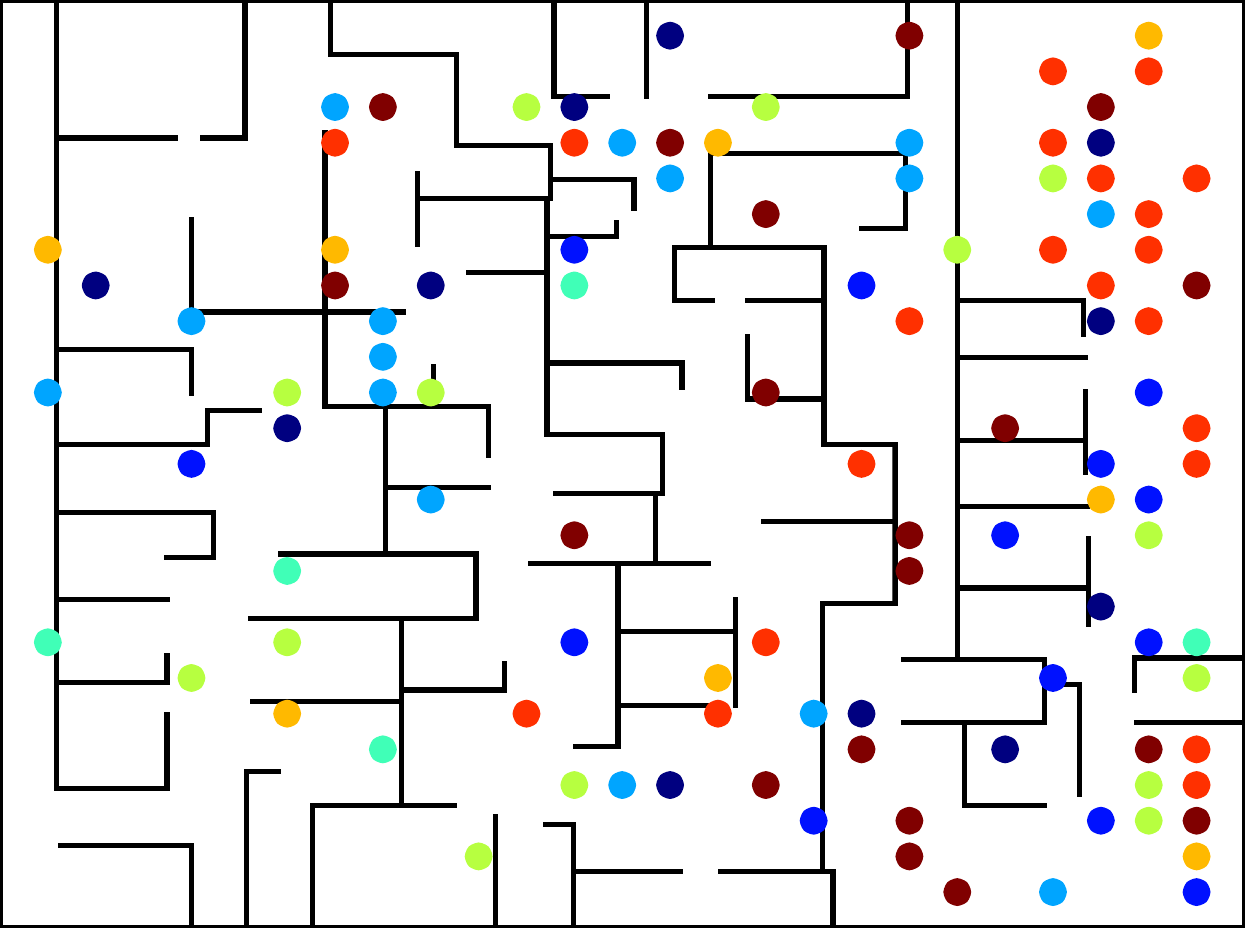}}\hfil
    \subfigure[\scriptsize Iter \# $20$k]{\includegraphics[width=0.192\textwidth]{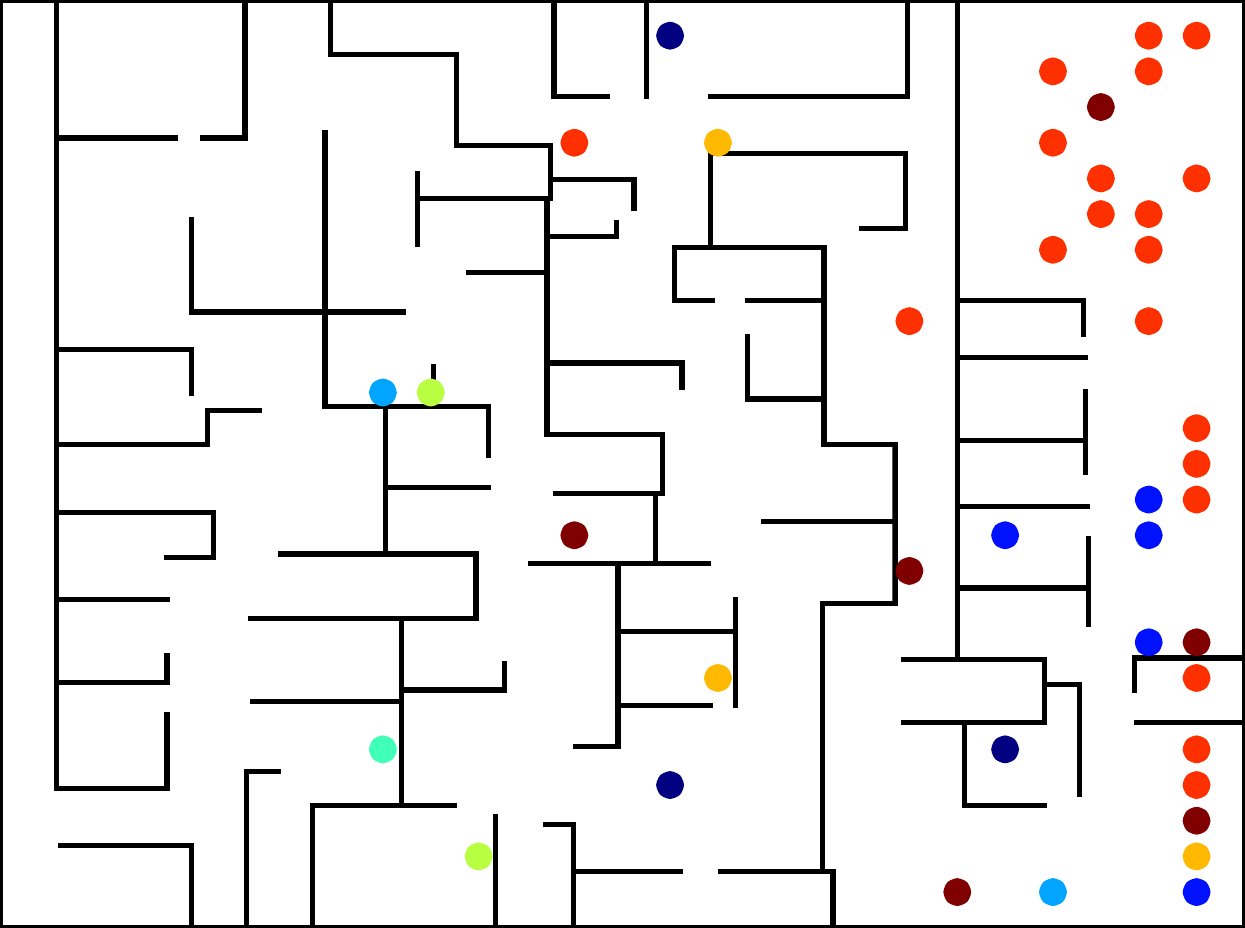}}\hfil
    \subfigure[\scriptsize Iter \# $30$k]{\includegraphics[width=0.192\textwidth]{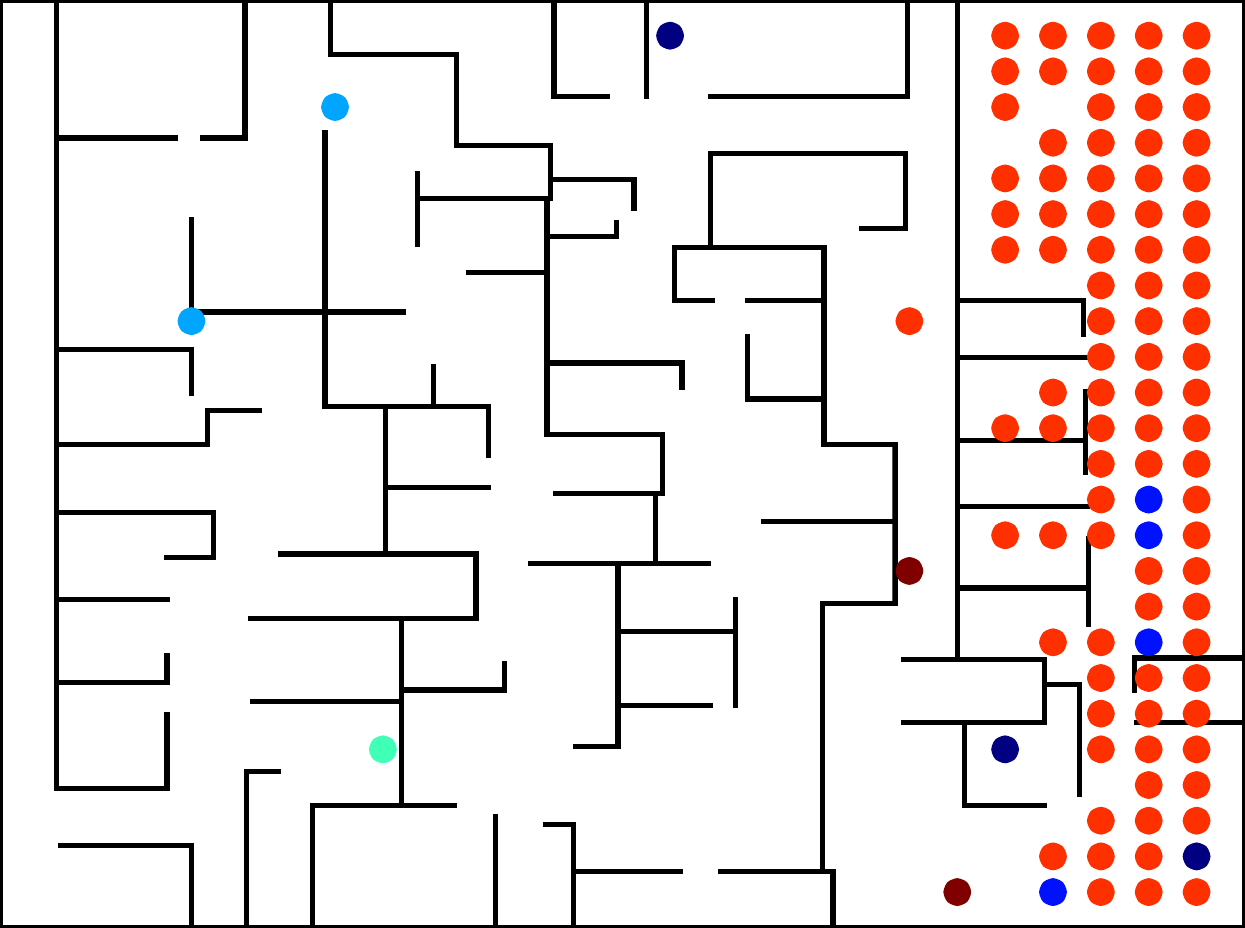}}\hfil
    \subfigure[\scriptsize Iter \# $70$k]{\includegraphics[width=0.192\textwidth]{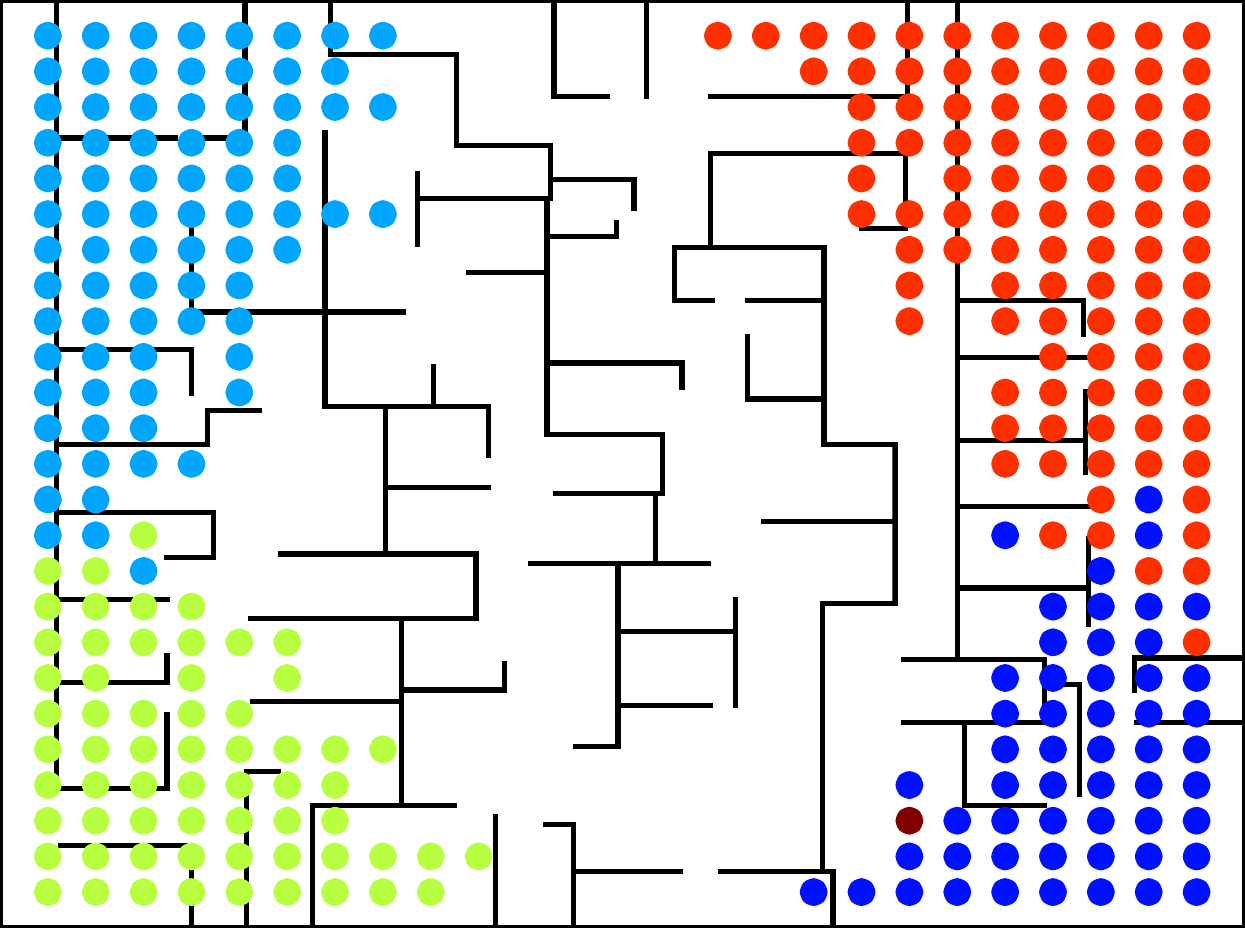}}\hfil
    \subfigure[\scriptsize Iter \# $140$k]{\includegraphics[width=0.192\textwidth]{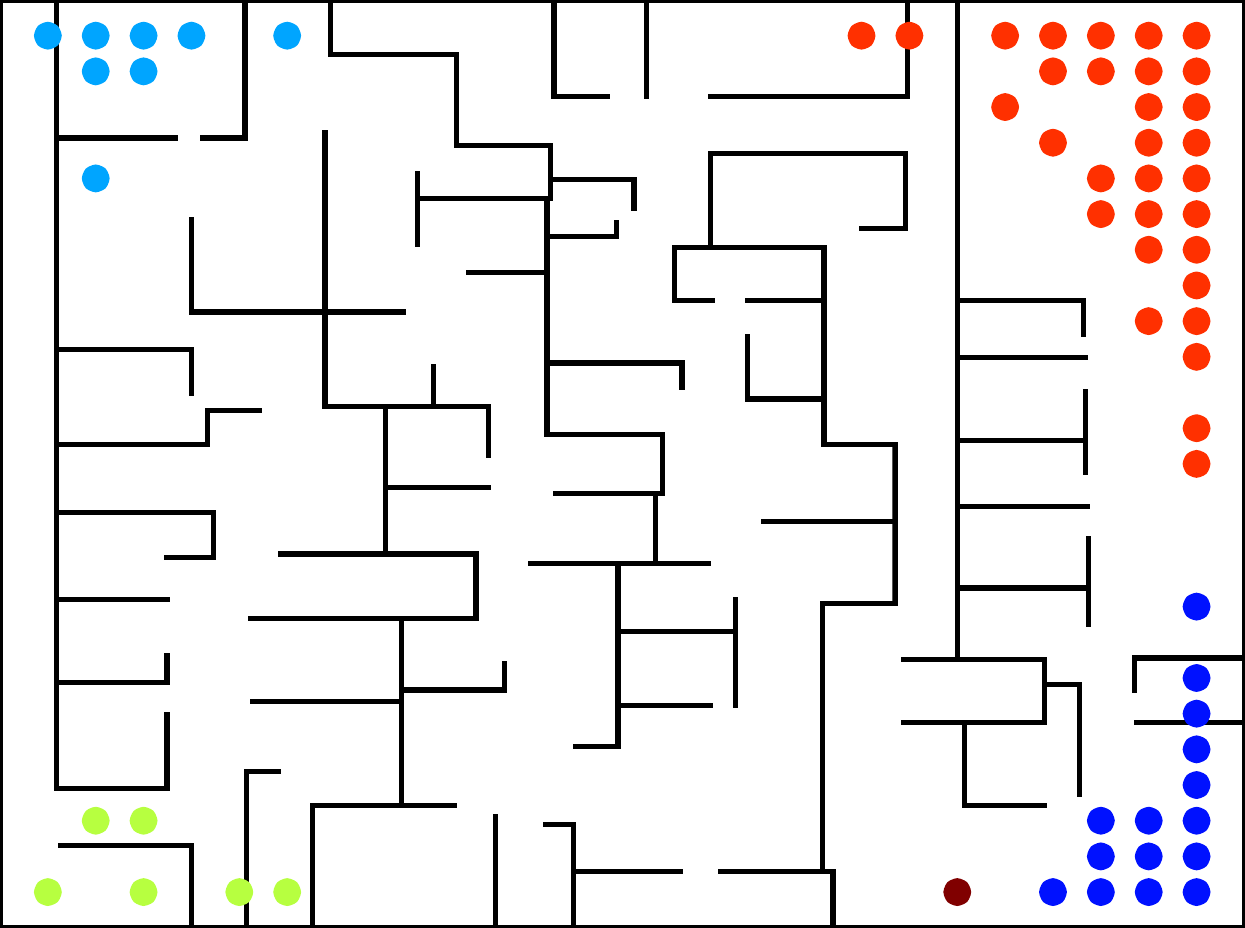}}\vspace{-0.65em}\\
    
    \subfigure[\scriptsize Iter \# $180$k]{\includegraphics[width=0.192\textwidth]{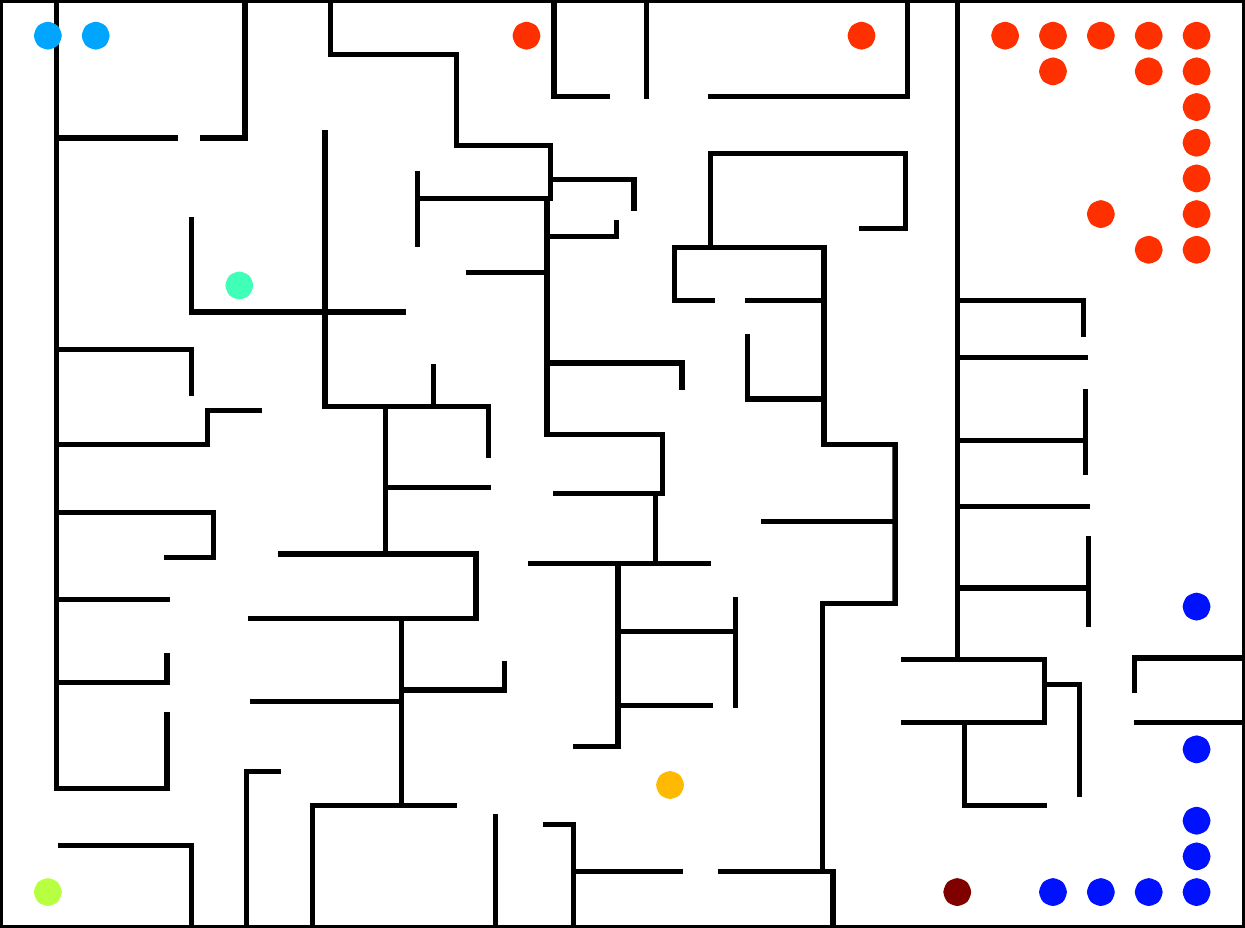}}\hfil
    \subfigure[\scriptsize Iter \# $250$k]{\includegraphics[width=0.192\textwidth]{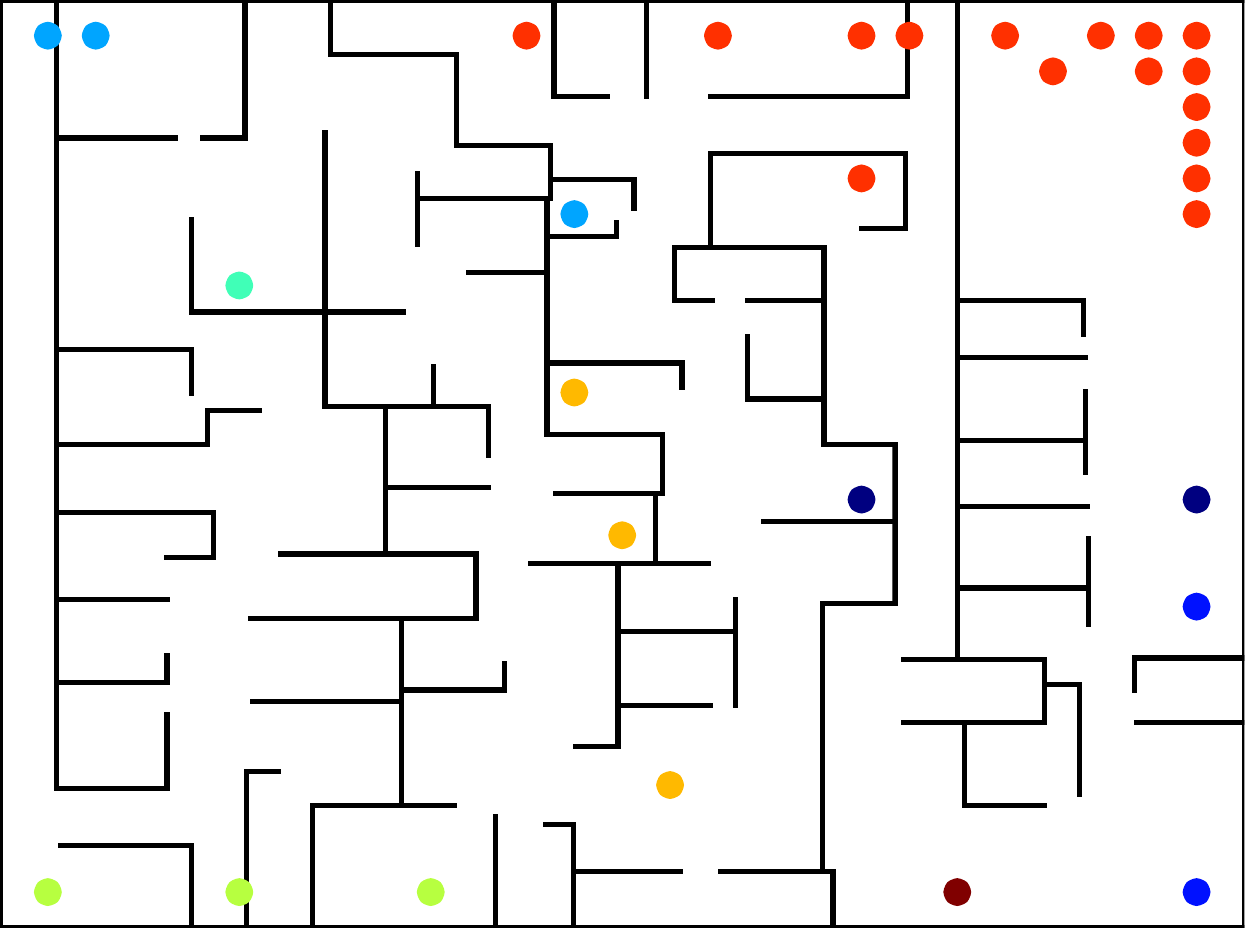}}\hfil
    \subfigure[\scriptsize Iter \# $300$k]{\includegraphics[width=0.192\textwidth]{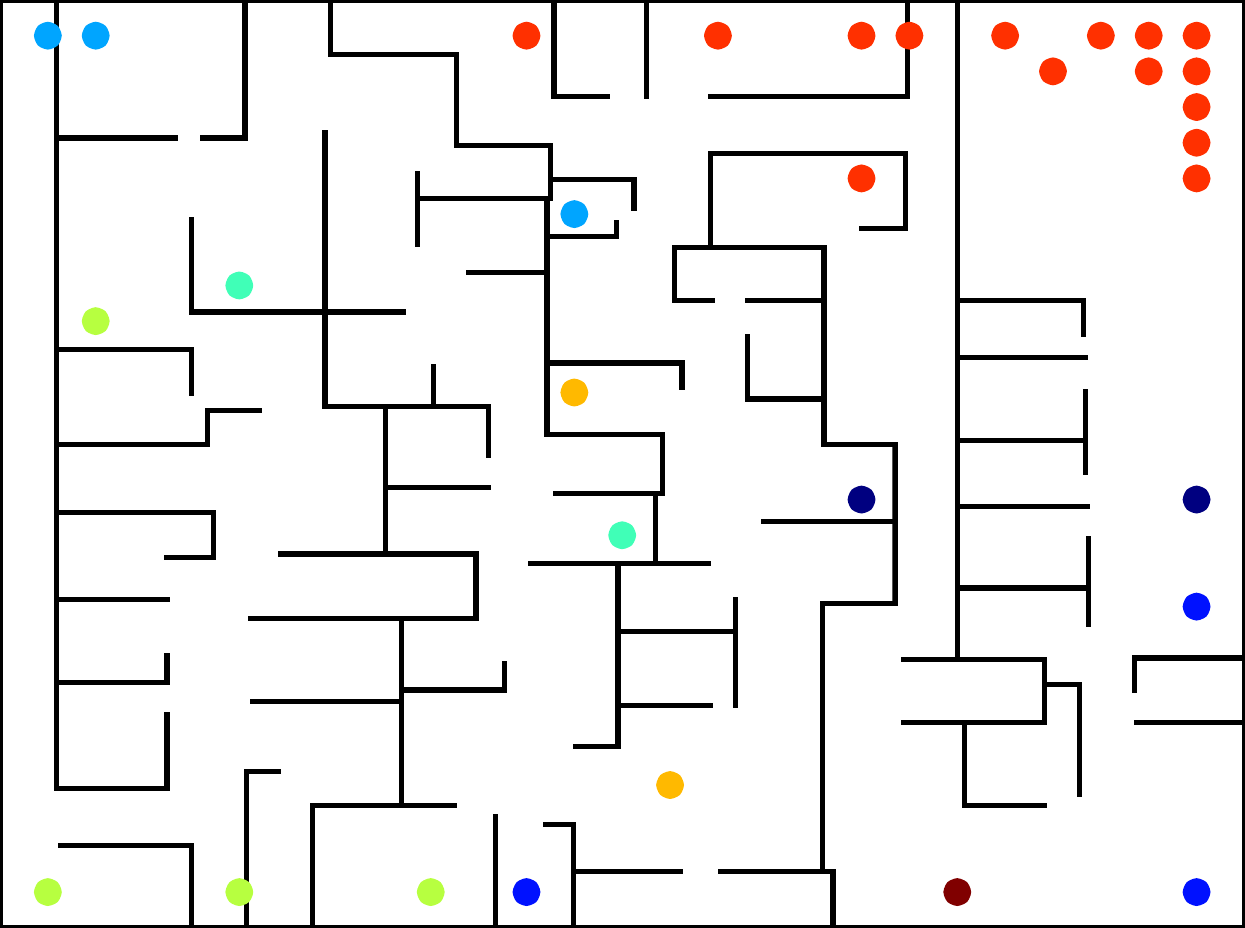}}\hfil
    \subfigure[\scriptsize Iter \# $500$k]{\includegraphics[width=0.192\textwidth]{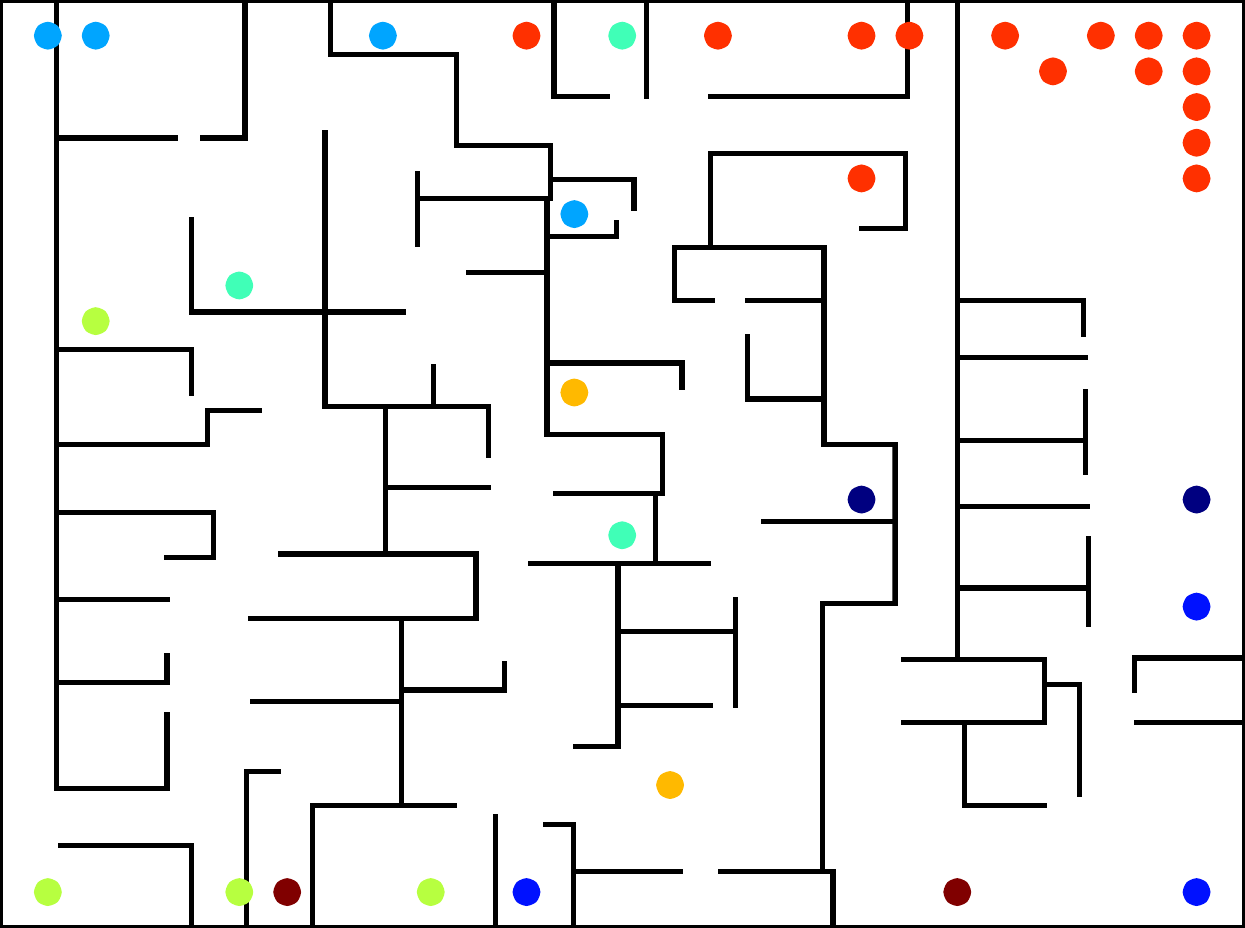}}\hfil
    \subfigure[\scriptsize Final]{\includegraphics[width=0.192\textwidth]{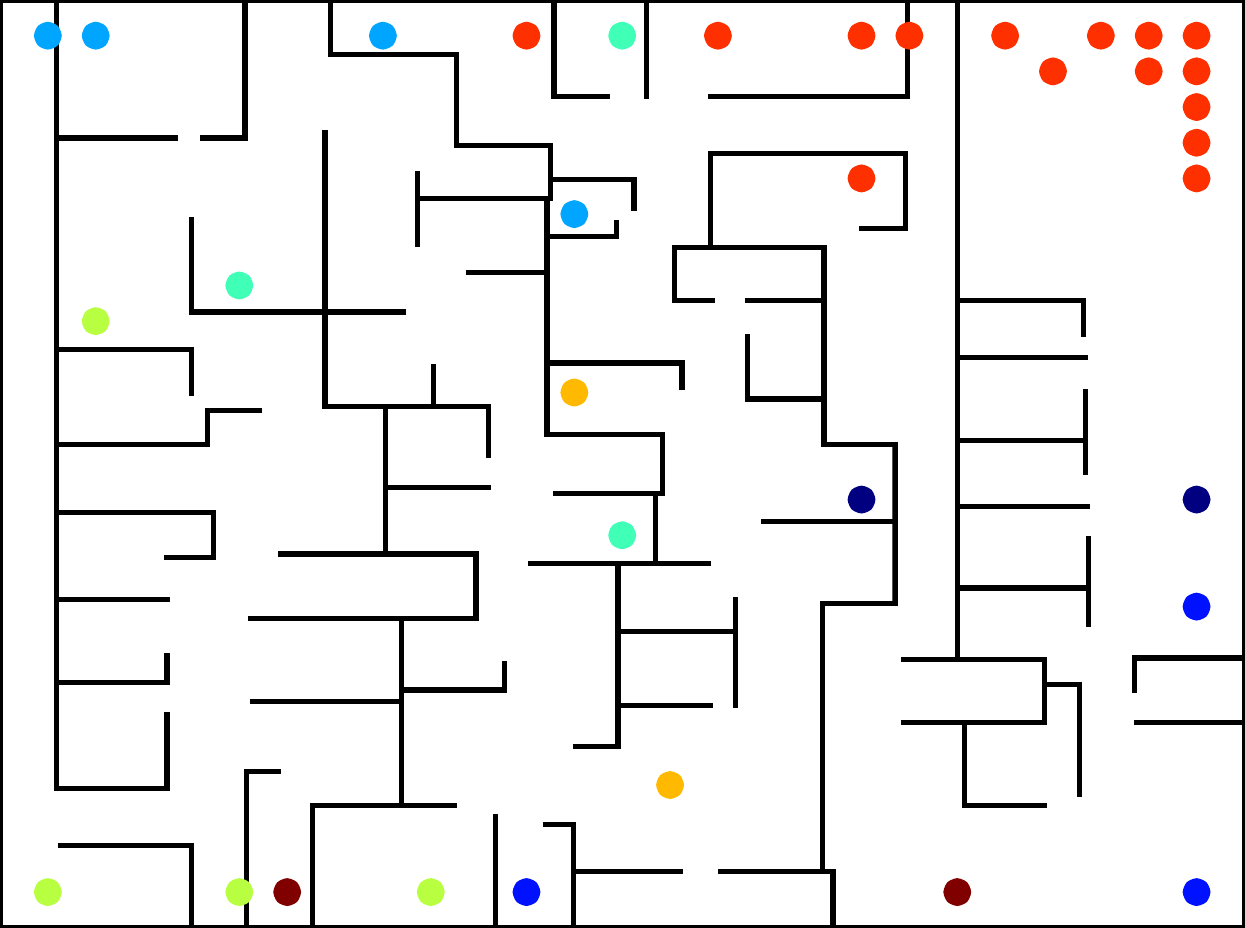}}\vspace{0.9em}\\

    {\scriptsize\bf Map 3}\\
    \subfigure[\scriptsize Iter \# $10$k]{\includegraphics[width=0.192\textwidth]{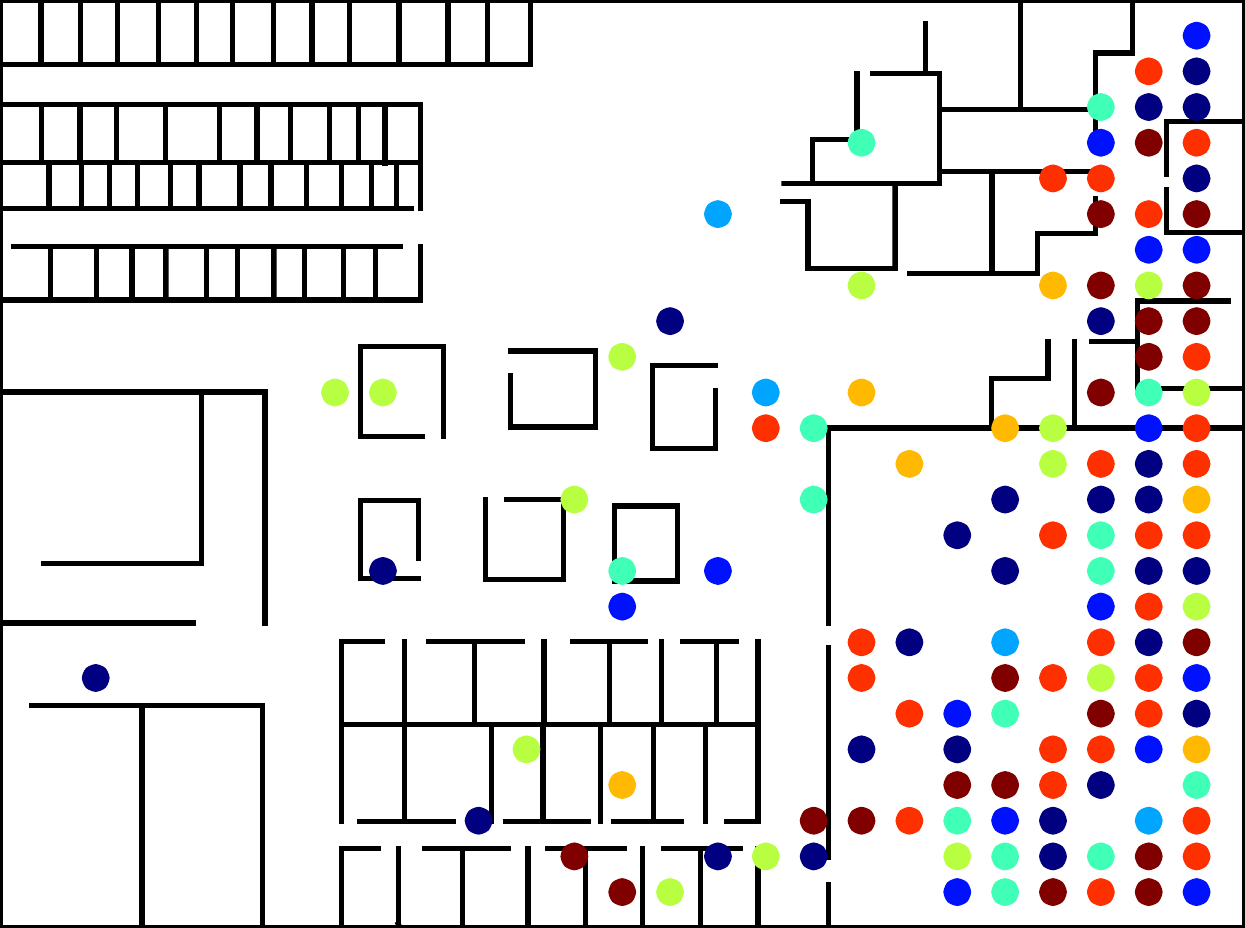}}\hfil
    \subfigure[\scriptsize Iter \# $20$k]{\includegraphics[width=0.192\textwidth]{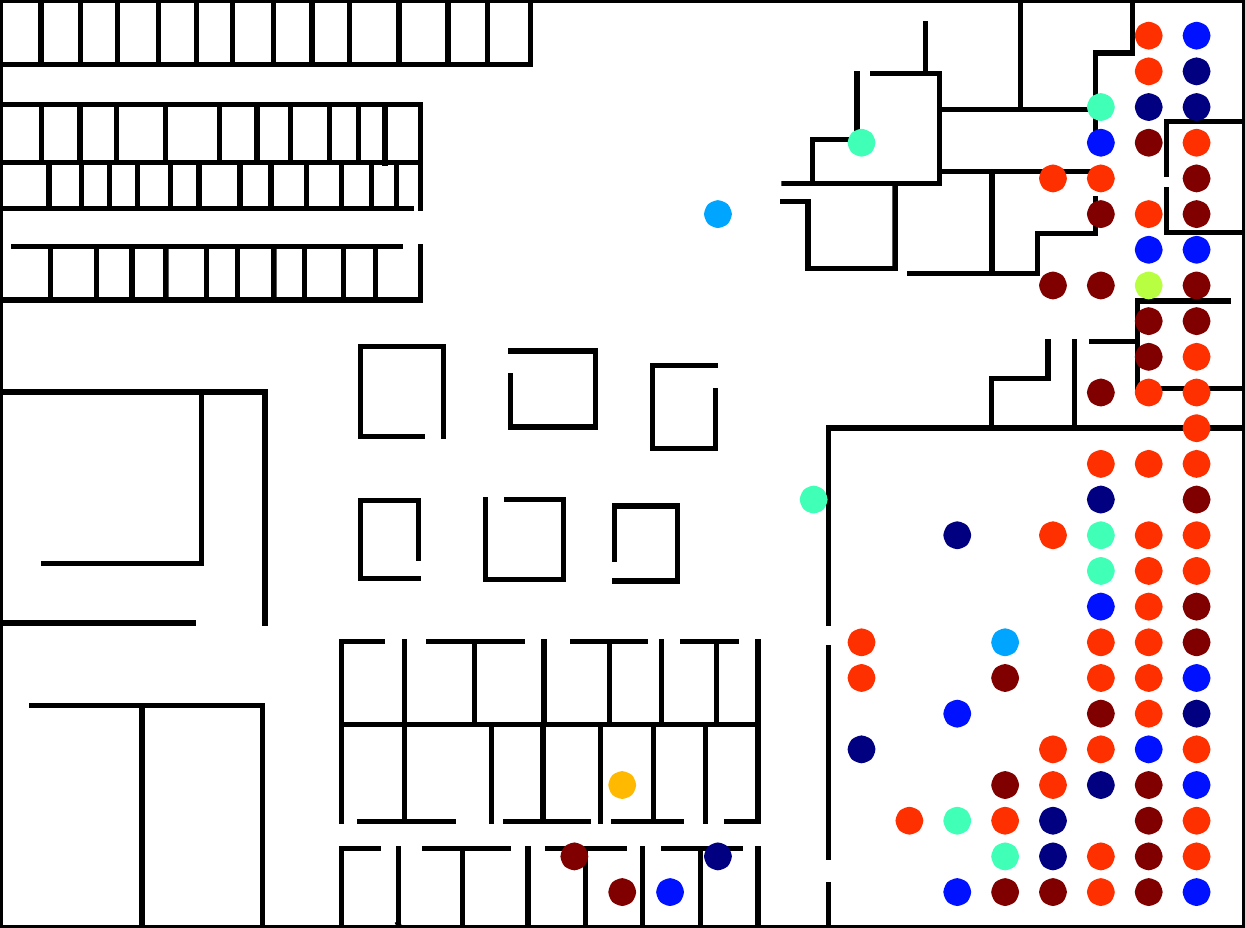}}\hfil
    \subfigure[\scriptsize Iter \# $30$k]{\includegraphics[width=0.192\textwidth]{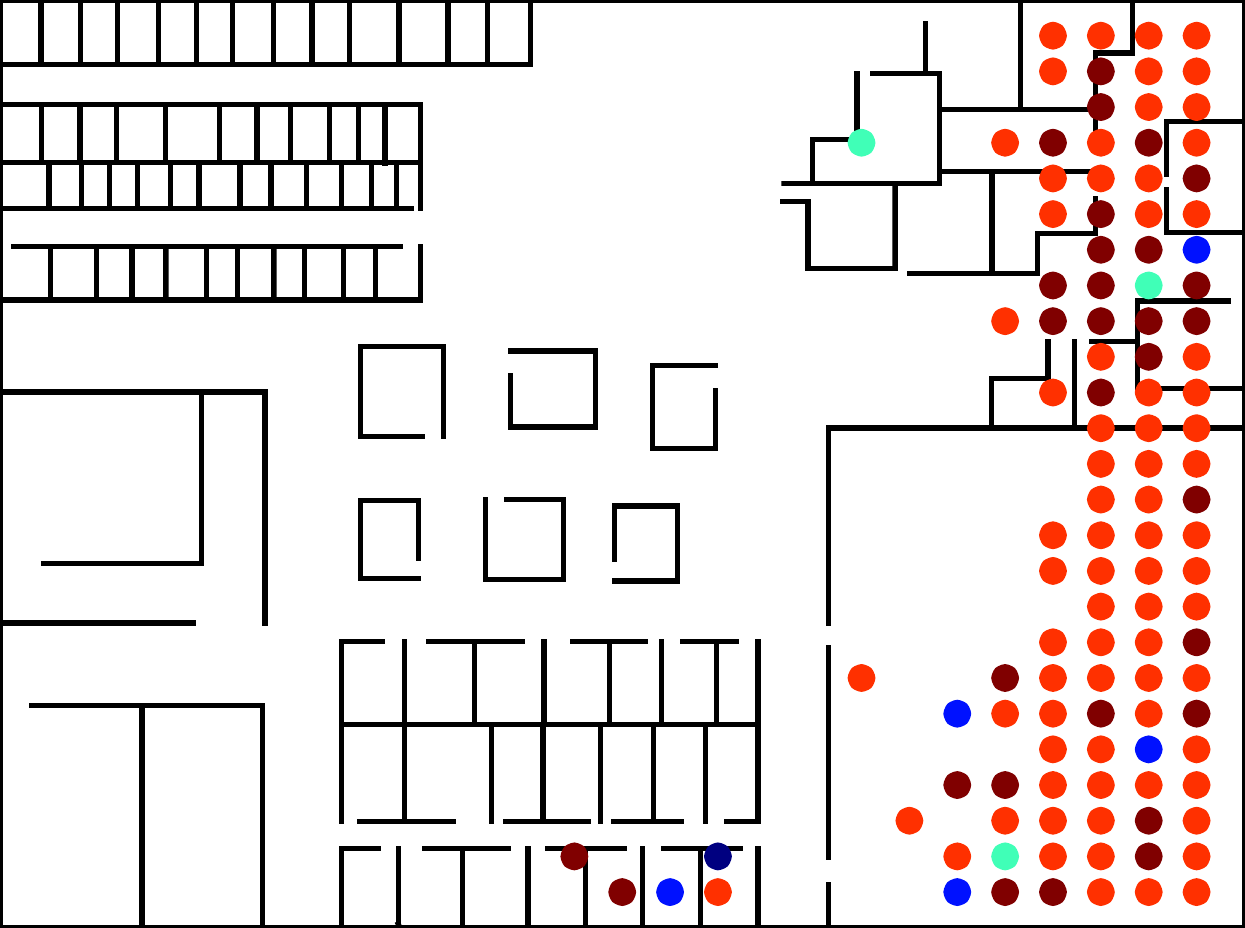}}\hfil
    \subfigure[\scriptsize Iter \# $70$k]{\includegraphics[width=0.192\textwidth]{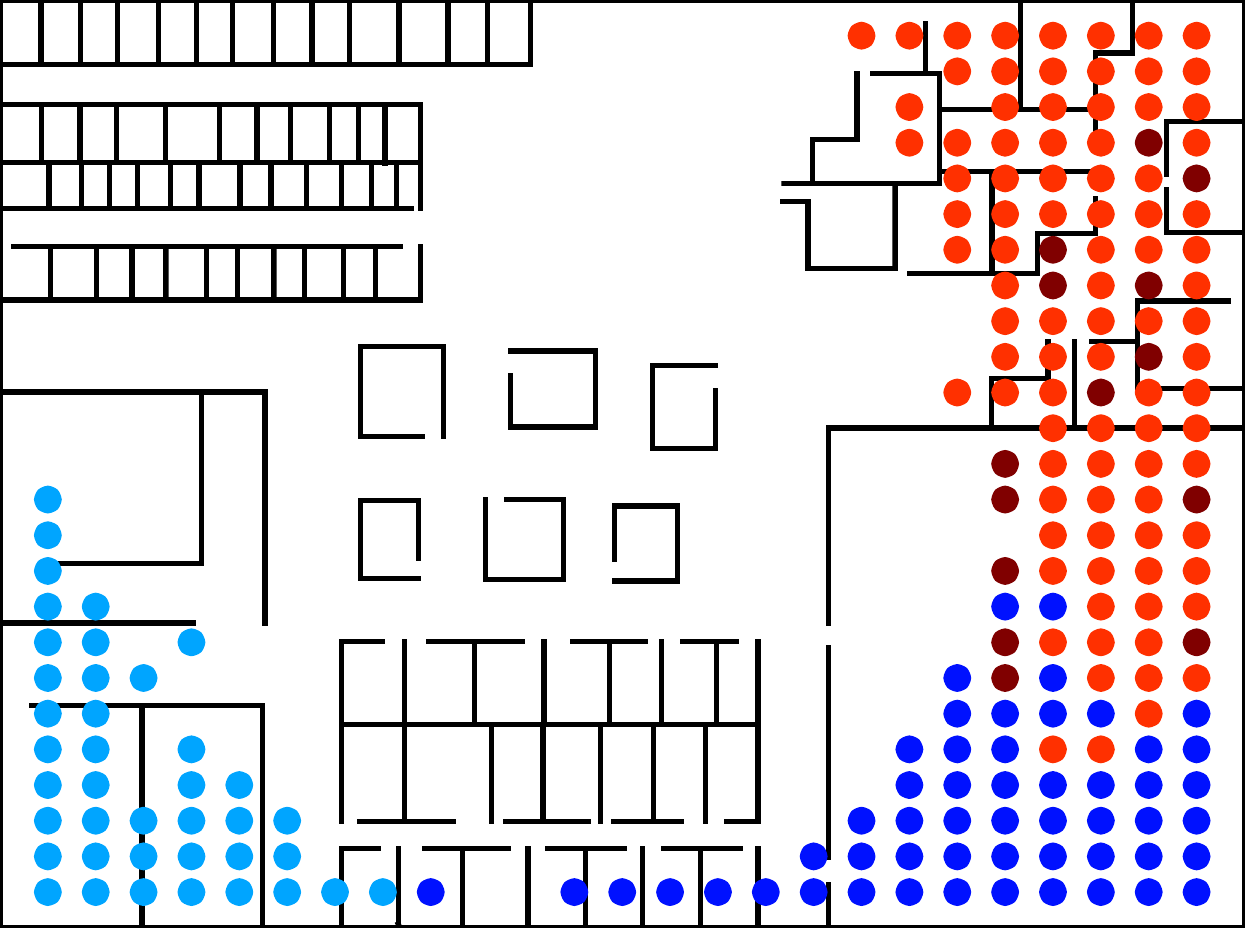}}\hfil
    \subfigure[\scriptsize Iter \# $140$k]{\includegraphics[width=0.192\textwidth]{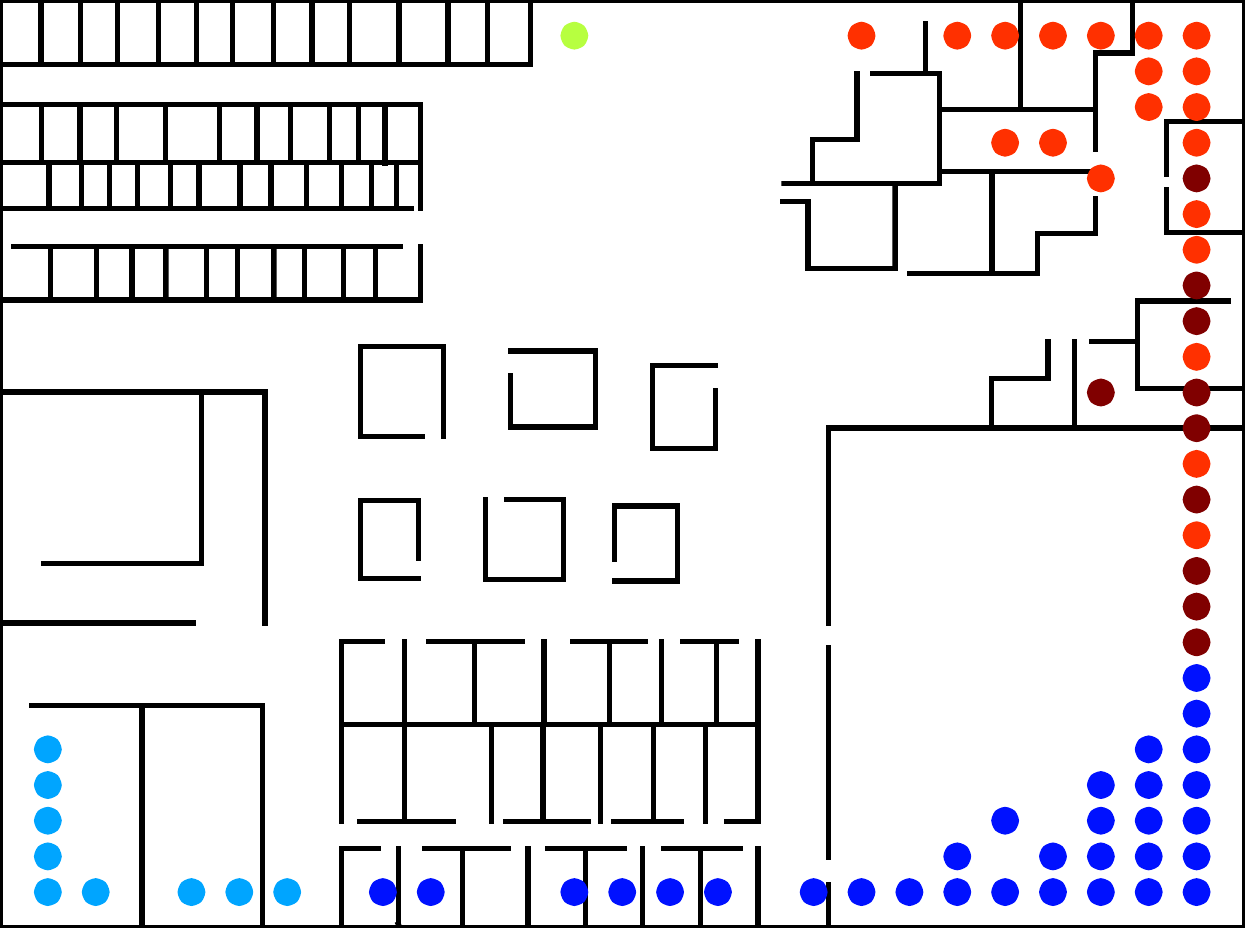}}\vspace{-0.65em}\\
    
    \subfigure[\scriptsize Iter \# $180$k]{\includegraphics[width=0.192\textwidth]{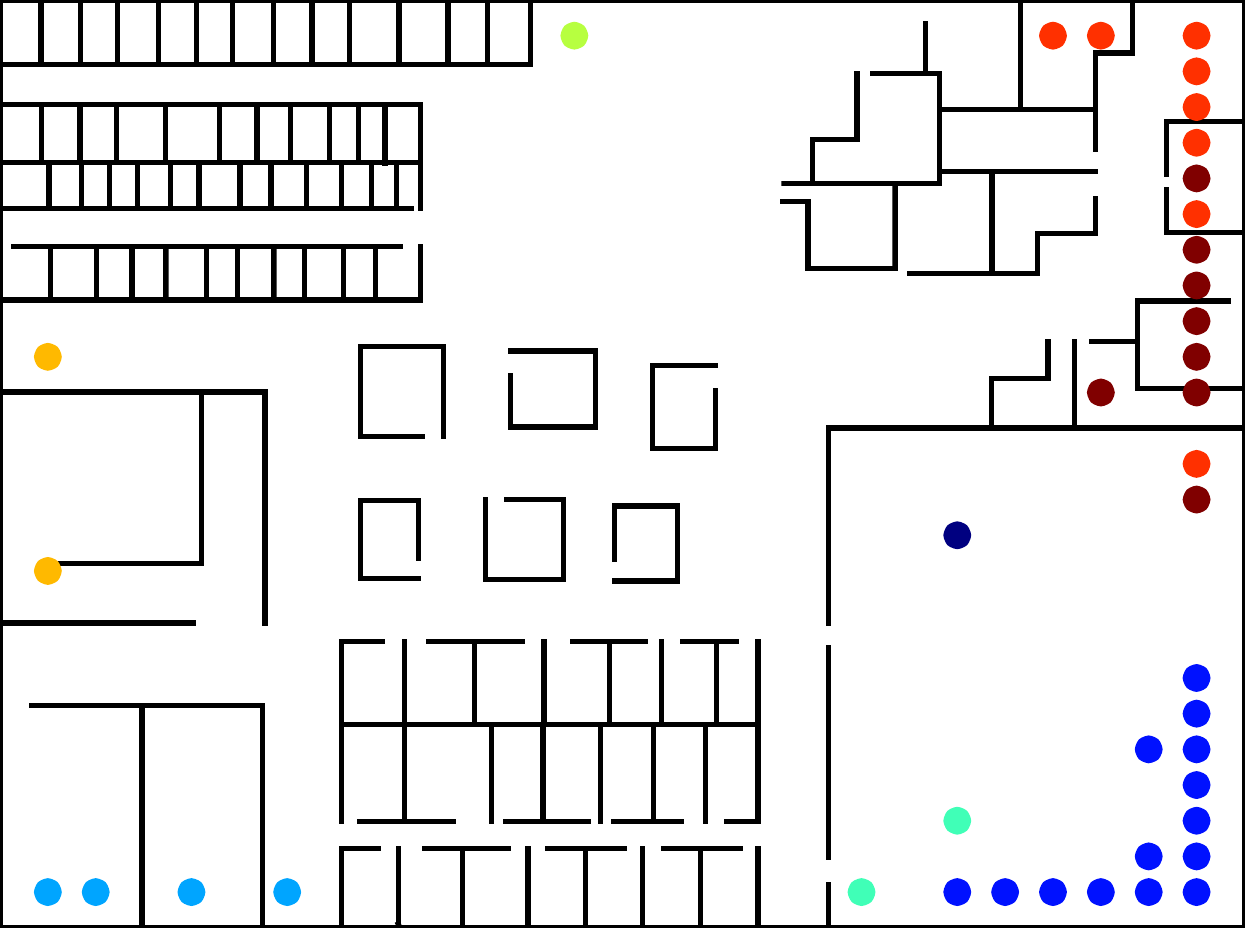}}\hfil
    \subfigure[\scriptsize Iter \# $250$k]{\includegraphics[width=0.192\textwidth]{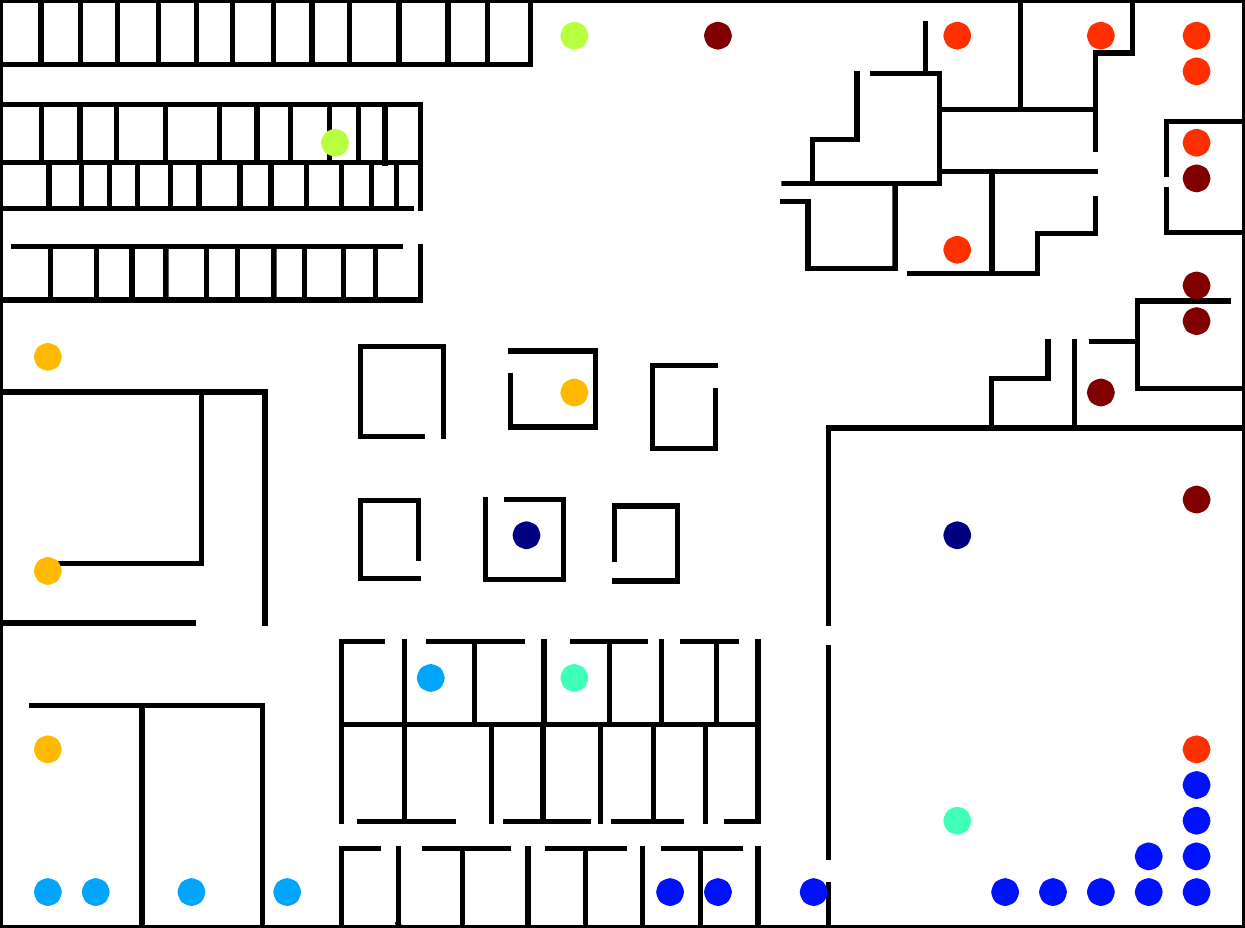}}\hfil
    \subfigure[\scriptsize Iter \# $300$k]{\includegraphics[width=0.192\textwidth]{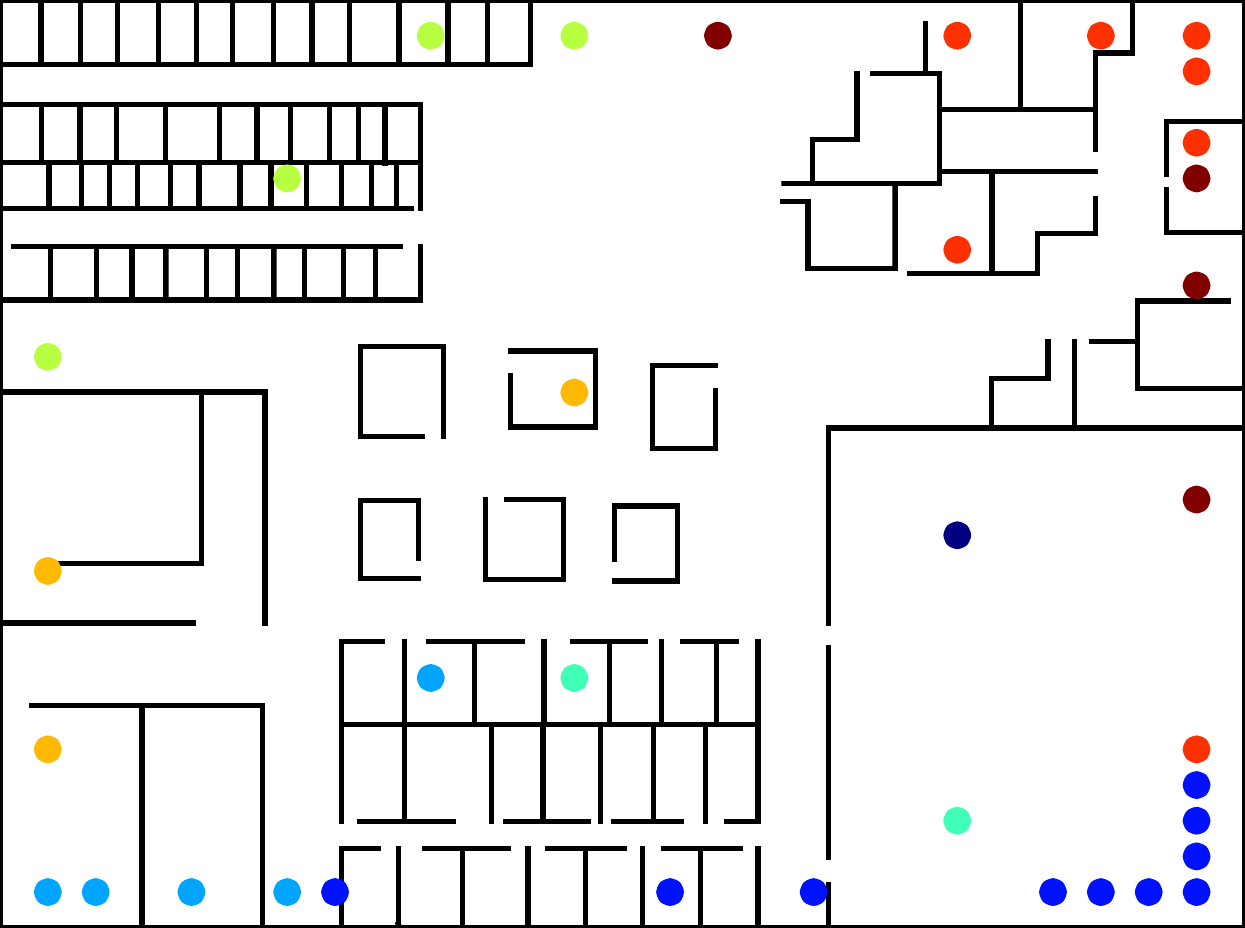}}\hfil
    \subfigure[\scriptsize Iter \# $500$k]{\includegraphics[width=0.192\textwidth]{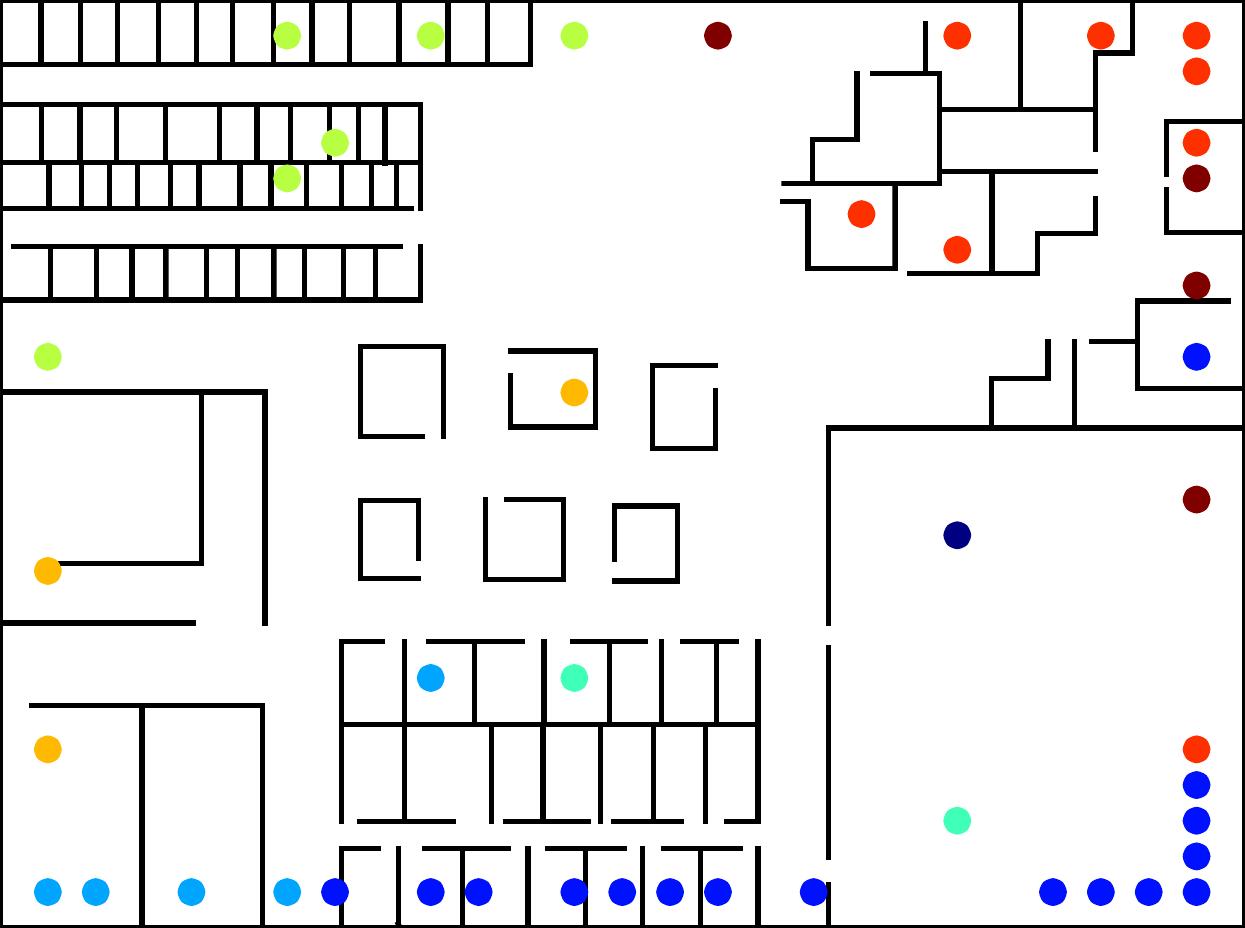}}\hfil
    \subfigure[\scriptsize Final]{\includegraphics[width=0.192\textwidth]{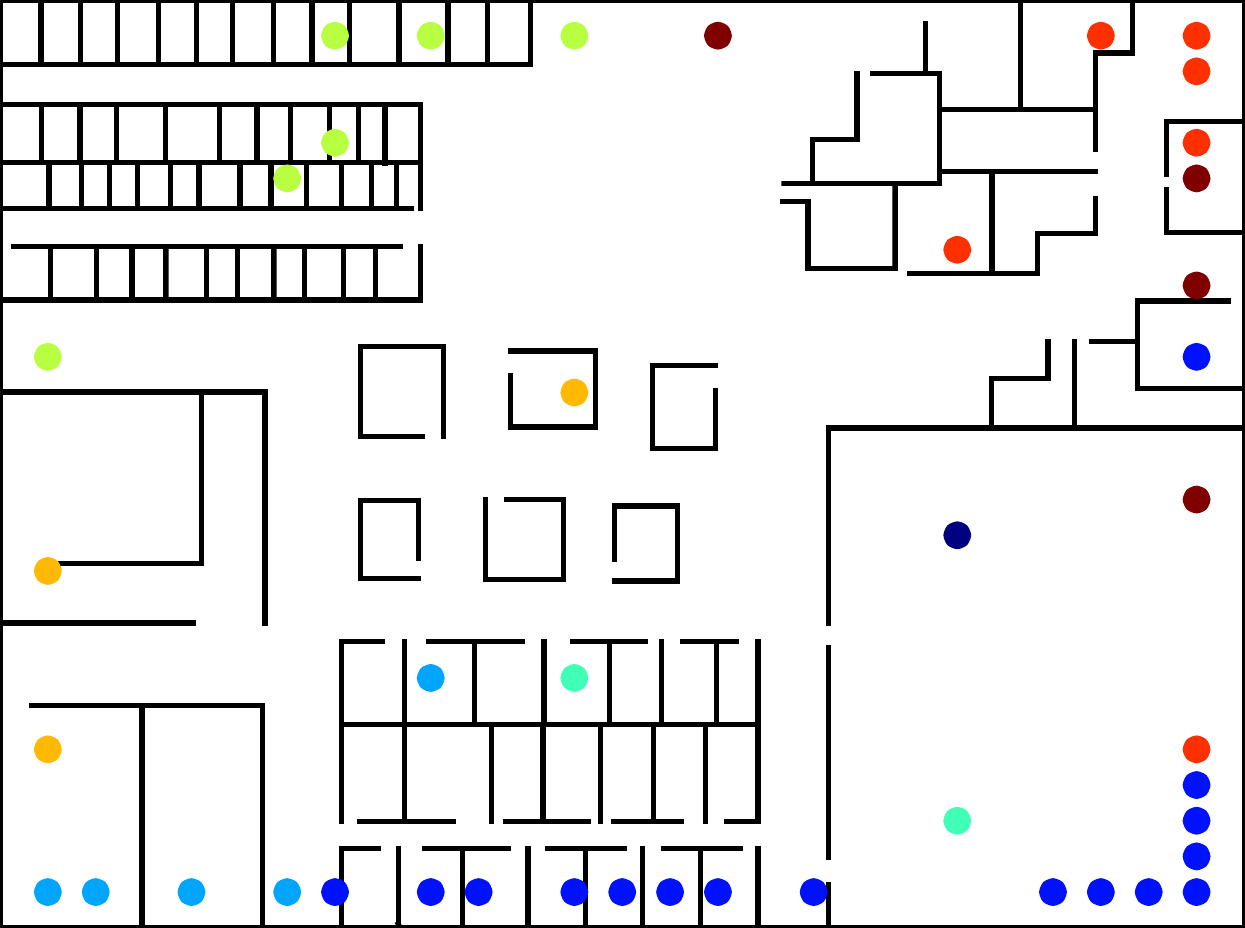}}

    \caption{The evolution of beacon distribution throughout training (with annealed regularization). The images depict a hard assignment, but the network is uncertain early in training, which explains the scarcity of beacons early on. The network quickly groups a large number of beacons and channel assignments along the edge of the map, but then gradually learns a sparse, diverse allocation, converging to a stable configuration around $200$k iterations.} \label{fig:map1-evolution}
\end{figure*}

Figure~\ref{fig:map1-evolution} shows the evolution of beacon distributions throughout training (when using an annealed regularizer). Note that the images depict a hard assignment, however the network reasons over high entropy placements early in training, which explains the initial sparsity. With each map, the network quickly clusters a large number of beacons by channel, and gradually learns to reduce the number of beacons while increasing channel diversity.

Next, we evaluate the ability of our method to automatically discover
successful placement and inference strategies for different
environmental conditions and constraints in
Table~\ref{tab:env-channels}. All results are on Map 1, with high
fixed regularization ($0.2$). We report results for a propagation
model with decreased attenuation at walls ($\beta = e^{-0.2}$), and
one with increased noise ($\sigma_z^2 = 2.5\times 10^{-4}$). We find
that our method adapts to these changes intuitively. Our method places
fewer beacons when the signal passes largely unattenuated through
walls and places more beacons when combating increased noise. We also
experiment with fewer ($4$) and more ($16$) available RF channels. As
expected, the availability of more channels allows our method to learn
a more accurate localization system. More broadly, these experiments
show that our approach can enable the automated design of location
awareness systems in diverse settings.

Finally, we evaluate the robustness of the joint placement and inference optimization with different random initializations. We repeat training on Map 1 (with fixed regularization of $0.04$) ten times,  and report the deviations in the error metrics and numbers of beacons placed in Table~\ref{tab:restart}.

\begin{table}[!t]
    \centering
    \setlength{\tabcolsep}{8.5pt}
    \caption{Modified Environment Settings}\label{tab:env-channels}
    \begin{tabularx}{1.0\linewidth}{l c c c}
        \toprule
        Scenario & Beacons & RMSE & RMSE (Worst-case)\\
        \midrule
        Original & $12$ & $0.0633$ & $0.1511$\\
        Low Attenuation & $\hphantom{0}8$ & $0.0426$ & $0.0910$\\
        High Noise & $27$ & $0.1400$ & $0.2907$\\
        Fewer Channels ($4$) & $11$ & $0.1290$ & $0.2581$ \\
        More Channels ($16$) & $12$ & $0.0397$ & $0.0910$ \\
        \bottomrule
    \end{tabularx}
\end{table}
\begin{table}[!t]
    \centering
    
    \caption{Training Stability over Multiple Runs}\label{tab:restart}
    \begin{tabularx}{0.9\linewidth}{l c c c c}
        \toprule
         & Mean & Std.\ Dev.\ & Min & Max\\
        \midrule
        RMSE & $0.057$ & $0.0019$ & $0.0541$ & $0.0596$\\
        Worst-case RMSE & $0.1365$ & $0.0039$ & $0.1289$ & $0.1413$\\
        Num.\ Beacons & $93.50$ & $11.17$ & $79$ & $119$\\
        \bottomrule
    \end{tabularx}
\end{table}

\section{Conclusion} \label{sec:conclusion}

We described a novel learning-based method capable of jointly optimizing beacon allocation (placement and channel assignment) and inference for localization tasks. Underlying our method is a neural network formulation of inference with an additional differentiable neural layer that encodes the beacon distribution. By jointly training the inference network and beacon layer, we automatically learn an optimal design of a location-awareness system for arbitrary environments. We evaluated our method for the task of RF-based localization and demonstrated its ability to consistently discover high-quality localization systems for a variety of environment layouts and propagation models, without expert supervision. Additionally, we presented a strategy that trades off the number of beacons placed and the achievable accuracy. While we describe our method in the context of localization, the approach generalizes to problems that involve estimating a broader class of spatial phenomena using sensor networks. A reference implementation of our algorithm is available on the project page at \url{http://ripl.ttic.edu/nbp}.

\section{Acknowledgements} \label{sec:acknowledgements}

We thank NVIDIA for the donation of Titan X GPUs used in this research. This work was supported in part by the National Science Foundation under Grant IIS-1638072.

\bibliographystyle{IEEEtranN}
{\small

}

\end{document}